\documentclass[review]{elsarticle}

\usepackage[ruled,vlined]{algorithm2e}
\usepackage{subcaption}
\usepackage{mathtools}
\usepackage{microtype}
\usepackage{longtable}
\usepackage{pdflscape}
\usepackage{setspace}
\usepackage{tabularx}
\usepackage{multirow}
\usepackage{ragged2e}
\usepackage{textcomp}
\usepackage{booktabs}
\usepackage{graphicx}
\usepackage{amsmath}
\usepackage{amssymb}
\usepackage{lscape}
\usepackage{lineno}
\usepackage{csquotes}
\usepackage[table]{xcolor}
\usepackage{caption}
\usepackage{array}
\usepackage{float}
\usepackage[breaklinks]{hyperref}
\hypersetup{
colorlinks=true,
linkcolor=blue,
filecolor=blue,
urlcolor=blue,
citecolor=blue,
}

\usepackage{geometry}
\modulolinenumbers[5]

\journal{arXiv}

\biboptions{numbers,sort&compress}
\begin{document}
\begin{frontmatter}
\title{AI Radiologist: Revolutionizing Liver Tissue Segmentation with Convolutional Neural Networks and a Clinician-Friendly GUI}
\author[QU_EE]{Ayman Al-Kababji}\corref{correspondingauthor}
\cortext[correspondingauthor]{Corresponding author}
\ead{aa1405810@qu.edu.qa}

\author[QU_EE]{Faycal Bensaali}
\ead{f.bensaali@qu.edu.qa}

\author[HMC]{Sarada Prasad Dakua}
\ead{SDakua@hamad.qa}

\author[UD]{Yassine Himeur}
\ead{yhimeur@ud.ac.ae}

\address[QU_EE]{Department of Electrical Engineering, Qatar University, Doha, Qatar}
\address[HMC]{Department of Surgery, Hamad Medical Corporation, Doha, Qatar}

\address[UD]{College of Engineering and Information Technology, University of Dubai, Dubai, UAE}

\begin{abstract}
Artificial Intelligence (AI) is a pervasive research topic, permeating various sectors and applications. In this study, we harness the power of AI, specifically convolutional neural networks (ConvNets), for segmenting liver tissues. It also focuses on developing a user-friendly graphical user interface (GUI) tool, \enquote{AI Radiologist}, enabling clinicians to effectively delineate different liver tissues (parenchyma, tumors, and vessels), thereby saving lives. This endeavor bridges the gap between academic research and practical, industrial applications. The GUI is a single-page application and is designed using the PyQt5 Python framework. The offline-available AI Radiologist resorts to three ConvNet models trained to segment all liver tissues. With respect to the Dice metric, the best liver ConvNet scores 98.16\%, the best tumor ConvNet scores 65.95\%, and the best vessel ConvNet scores 51.94\%.  It outputs 2D slices of the liver, tumors, and vessels, along with 3D interpolations in .obj and .mtl formats, which can be visualized/printed using any 3D-compatible software. Thus, the AI Radiologist offers a convenient tool for clinicians to perform liver tissue segmentation and 3D interpolation employing state-of-the-art models for tissues segmentation. With the provided capacity to select the volumes and pre-trained models, the clinicians can leave the rest to the AI Radiologist.
\end{abstract}


\end{frontmatter}

\section{Introduction}\label{sec1}

Every year, diseases associated with the liver lead to approximately two million deaths globally~\cite{Asrani2019}. About half of them are complications arising from liver cirrhosis, and the other half due to hepatitis and hepatocellular carcinoma (HCC)~\cite{Asrani2019}. Furthermore, the liver frequently acts as a primary site for metastases from other nearby organs, including the rectum, colon, stomach, pancreas, breasts, esophagus, and lungs~\cite{Surgery2020}. Because tumors can originate in the liver or spread from adjacent areas, examining the liver and its abnormalities is critical during the initial stages of tumor staging~\cite{Christ2017}. Systematic screening for liver diseases can substantially lower the risk of death~\cite{NASIRI_2020}. Additionally, the early identification and accurate segmentation of liver tumors are crucial in enabling healthcare providers to develop more targeted and effective treatment strategies.

The liver possesses two distinctive characteristics: i) its capability to function even when reduced to a portion of its original volume, thereby maintaining general health (however, if less than 20\% of the original liver volume remains post-surgery, it can lead to liver failure~\cite{Pagano2014}); and ii) its regenerative capacity, which enables it to recover back to its original volume over time. Despite these traits, hepatectomy operations carry a degree of risk. A notable 4.7\% of patients, unfortunately, succumb to the surgery, either during their hospital stay (2.9\%) or within 30 days post-operation (1.8\%), rendering these cases unsuccessful~\cite{Lee2016}. Furthermore, complications arising from liver resection operations affect nearly 35\% of patients, as reported in~\cite{Lee2016}, underscoring the need for better surgical planning and post-operative care.

In such complex scenarios, computer-aided liver tumor detection (CALTD) systems can provide valuable assistance to clinicians and surgeons. CALTD systems are used by medical professionals to enhance their interpretation of medical images acquired through diverse imaging modalities such as computerized tomography (CT) scans, magnetic resonance imaging (MRI), and ultrasound (US). Traditionally, these systems utilize statistical-based algorithms executed on images derived from the aforementioned modalities (either independently or in conjunction with other relevant diagnostic data and biomarkers)~\cite{Petrick2013}. These systems offer an algorithmic perspective, providing valuable supplemental insights for medical personnel.

Prominent architectures, such as the U-Net~\cite{Ronneberger2015}, have gained substantial traction in the biomedical field, particularly in the areas of organ and tissue segmentation. Integrating convolutional neural networks (ConvNets) with CALTD systems can create powerful tools that clinicians and surgeons can utilize for preparatory procedures. These machine learning (ML)-powered CALTD systems can then be deployed to segment all the liver tissues, thereby facilitating effective surgical pre-planning. By using such systems, healthcare providers can benefit from enhanced visualization and increased precision, potentially improving surgical outcomes and patient recovery times.

\begin{figure}[!ht]
    \centering
    \includegraphics[width=0.9\linewidth]{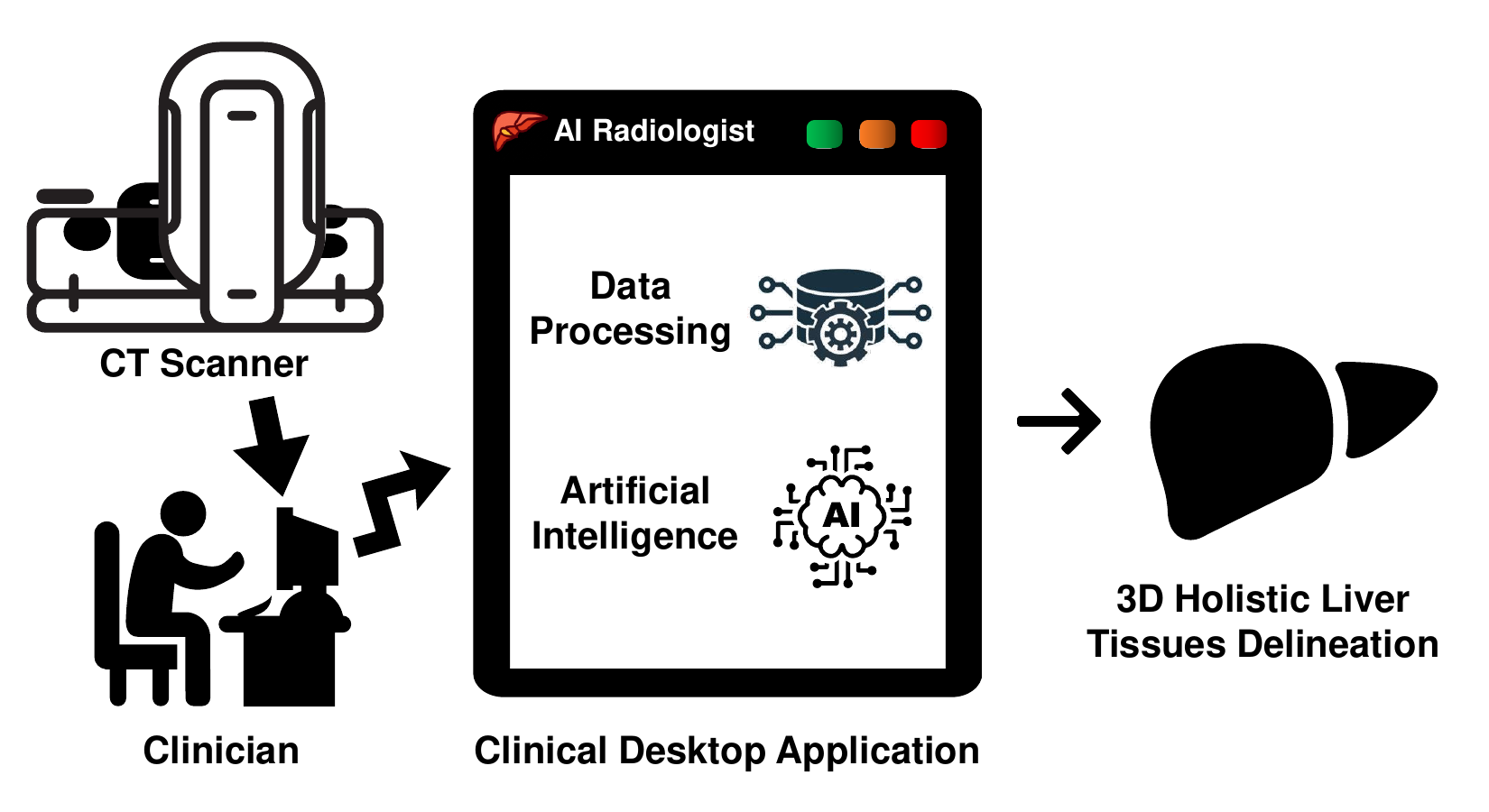}
    \caption{System Schema.}
    \label{fig:system_schema}
\end{figure}

This paper explores the deep learning technique's algorithmic implementation. Moreover, it focuses on developing the AI Radiologist system, as illustrated by Figure~\ref{fig:system_schema}. The AI Radiologist system empowers clinicians to execute detailed segmentation of liver tissues (parenchyma, tumors, and vessels), followed by a 3D object interpolation of the segmented tissues.
We aim to develop an intuitive tool that significantly minimizes the learning curve associated with using the AI Radiologist system. The system takes a CT scan of \enquote{Neuroimaging Informatics Technology Initiative (NIfTI)} format as an input and generates the automatic segmentation for each tissue alongside a 3D liver object as an output.
We chose the output to be formatted in .obj and .mtl, given their ability to include multiple objects. Furthermore, these file types are frequently used with 3D printers for multi-object printing, providing another level of practical utility. The AI Radiologist system is thus designed not only to enhance clinical insights and decision-making but also to facilitate potential applications such as 3D printing, aiding in the visual and tactile understanding of complex liver structures.

This paper elucidates the application of the AI Radiologist system in segmenting liver tissues, including tumors and vessels, from medical CT images. The key contributions can be encapsulated as follows:

\begin{itemize}
\item Introducing AI Radiologist, an offline desktop application powered by AI, equipped with pre-trained models for the liver tissue segmentation challenge. This application delivers robust computational power while requiring minimal user interaction and expertise.

\item Addressing the segmentation of all tissues within the liver, this study stands out as one of the few that deliver results surpassing the state-of-the-art in liver segmentation and offering comparable outcomes for tumor and vessel segmentation.

\item Providing an in-depth explanation of the methodology for creating multi-tissue objects, which are the outputs of our AI Radiologist system. These objects are readily available for immediate 3D printing, demonstrating the practical applicability of the system.

\item Demonstrating the effectiveness of the ConvNet models through a 5-fold cross-validation method across several performance metrics such as intersection-over-union (IoU), Dice similarity coefficient (DICE), and Hausdorff distance (HD95).

\item Simplifying the clinical workflow by allowing clinicians to efficiently select and process CT scans through the AI Radiologist, eliminating the need for in-depth technical knowledge.

\item Setting a foundation for future technological advancements in medical diagnostics and proposing enhancements to expand the tool’s capabilities, including compatibility with additional file formats and improved segmentation results for broader clinical applications.

\end{itemize}

The structure of the remainder of this paper is as follows: Section~\ref{sec2} delves into the relevant studies associated with the utilization of desktop application systems for clinical segmentation tasks, alongside the integration of ML and ConvNets in these challenges. In Section~\ref{sec3}, we outline the adopted methodology for achieving both the ConvNets results and the creation of the AI Radiologist application. Section~\ref{sec4} furnishes the ConvNets results and provides a demonstration of the developed AI Radiologist application. Finally, we conclude in Section~\ref{sec5}.

\section{Related Work}\label{sec2}
Given the myriad benefits of surgical pre-planning, a host of technological solutions have emerged to equip surgeons with the best possible surgical planning tools. We delve into two different aspects of related works, the former being about the different automated ML techniques deployed for any liver's tissues segmentation, and the latter focusing on the solutions' deployment for experts to use even if unrelated to the liver. 

\subsection{Liver Tissues Segmentation Solutions}
Researchers dealt with the liver tissue segmentation challenge using different approaches based on historical and current algorithmic advancements. In this section, we mention different approaches and the prominent ones to be compared within the results section, attempting to organize the methodologies through different periods.

\subsubsection{Unsupervised and Supervised Miscellaneous Models}
Among various liver tissue segmentation tasks, the segmentation of liver parenchyma has notably been addressed using a range of unsupervised machine learning algorithms, which generally achieve commendable results. This effectiveness partly stems from the liver's nature as a large, single organ where pixel intensities are consistent within the same volume. Prominent examples of such algorithms are k-means and fuzzy c-means (FCM). Various implementations of k-means and FCM have been explored in~\cite{Huang2018,Cai2019,Ivashchenko2020} for primary segmentation tasks, and in specific instances like in~\cite{Sayed2016}, FCM's efficacy is enhanced through integration with grey wolf optimization algorithm, particle swarm optimization, and neutrosophic sets. In contrast, tumor segmentation has attracted fewer studies, but those that exist also frequently employ FCM variants, as noted in~\cite{Das2019,Anter2019}. Conversely, the use of unsupervised algorithms for vessel segmentation is relatively uncommon and is mainly discussed in~\cite{Zhang2018vessels}, where k-means clustering forms the basis for subsequent enhancements like Jerman’s vesselness filter and an advanced fuzzy connectedness (FC) algorithm. Despite their utility, the reliance on unsupervised methods for segmentation can yield results that are not always optimal or reliable for critical medical decisions. Consequently, many researchers advocate for supervised learning approaches as more robust solutions for segmenting liver tissues.

Naturally, given the biomedical field's criticality, supervised techniques were much preferred due to their extraordinary performance compared to unsupervised ones. A few examples would comprise random forests (RF)s, which have been deployed for numerous tasks related to liver segmentation as in~\cite{Lombaert2014,Norajitra2017, Treilhard2017}, and adaptive boosting (AdaBoost) as in~\cite{Zhang2015,Zheng2017automatic}. For tumor segmentation, only ConvNet-based solutions were evident in previous literature, and supervised ML-based solutions are negligible. Contrastingly, in~\cite{Zeng2016}, an extreme learning machine (ELM) is used after heavy usage of filters (anisotropic, Sato, Frangi, offset medialness, and strain energy) to extract vasculature features for vessels segmentation.

Authors in~\cite{Treilhard2017} created the sole study utilizing supervised-based ML that segments all the liver's tissues, i.e., liver parenchyma, tumors, and vessels, where cascaded RFs is the chosen tool to take on that multi-tissue challenge in that study.

Even though they generate better results, they lack performance for the biomedical field's criticality level. This motivated researchers to use the more robust and complex models attempting to learn the pattern of inter-human livers. Thus, the use of ConvNets and generative adversarial networks (GANs) are deeply investigated as they show more promising results that have never been witnessed before. 

\subsubsection{Fully Convolutional Networks (FCN) Models}
FCN-based solutions are the natural choice for these types of challenges where the data is of an imagery nature, having both x and y dimensions (i.e., 2D). They fall under the supervised ML schema. For the liver segmentation task,~\cite{Yuan2017,Zheng2017,Tian2018,Qin2018,Jansen2019} opted to use 2D FCNs with some variations and with the help of other algorithms, such as using 3D deformable model optimization (3D DMO) and negative matrix factorization (NMF) in~\cite{Zheng2017} or using simple linear iterative clustering (SLIC) algorithm to generate superpixels in~\cite{Qin2018}. This aided the performance of such FCNs but also required more computations outside the devised ML algorithm to segment the liver. Some of the aforementioned studies have also used the same FCN for tumor segmentation within, such as in~\cite{Zheng2017,Tian2018}, while researchers in~\cite{Christ2016,Christ2017,Yuan2017} opted to use a different FCN for the tumor segmentation.

On one hand, researchers such as~\cite{Sun2017}, opt to input the CT image full-size into a single 2D FCN on single-phase CT or 2D multichannel FCN (MC-FCN) to segment tumors within the liver, but they do not train the networks from scratch, on the contrary, those networks are initially trained on the liver segmentation task for faster convergence when trained for tumor segmentation. On the other hand, others like~\cite{Li2015} divide the CT scans further into patches to accommodate the scattered nature of tumors and their relatively small size with respect to the liver in many cases. Also, when patching the images, the model has to deal with fewer irrelevant pixels/voxels within a frame.

The segmentation of vessel tissues predominantly utilizes supervised ConvNet-based machine learning algorithms. In~\cite{Mishra2019}, a 2D deeply supervised network (DSN) that incorporates VGG-16 architecture is employed for segmenting liver vessels from ultrasound images. Another approach by~\cite{Kitrungrotsakul2019} introduces a 2D VesselNet, which integrates three DenseNets to segment orthogonal patches across three planar views, subsequently combining vesselness probability maps to produce a final segmentation mask.

Significantly, there has been an interest in exploring 3D fully convolutional networks (FCNs) to leverage the volumetric data present in CT scans. One pioneering work by~\cite{Lu2016} uses a 3D FCN for liver segmentation, enhanced by graph cut (GC) techniques. Following this, studies by~\cite{Hu2016,Hu2017} adapt the same 3D FCN architecture, enlarging kernel sizes and altering activation functions to optimize the segmentation output through a global energy function. Moreover, ~\cite{Dou2016,Dou2017} introduce a deep supervision mechanism into their 3D FCN, paired with a 3D conditional random field (CRF) for refining segmentation results. In another development, ~\cite{Gibson2017} advances a 3D FCN based on the Dense V-Net framework, which was further expanded to multi-organ segmentation, including the liver, in~\cite{Gibson2018}. Likewise, the 3D anisotropic hybrid network (3D AH-Net) detailed in~\cite{Liu2018} innovatively transforms 2D trained weights into a 3D format, aiding both parenchyma and tumor segmentation. While the shift to 3D FCNs aims to capture complex volumetric patterns, these models often come with high memory and computational demands, making them challenging to implement in clinical settings where resources are limited or rapid responses are necessary.

To alleviate the computational burden of 3D FCNs while still incorporating volumetric data, the strategy of using 2.5D inputs has been devised. This approach focuses segmentation on a single slice per forward pass but includes its adjacent slices in the input, effectively using $2(k-1)$ neighboring slices to provide additional spatial context. This setup allows for the generation of a segmentation mask solely for the central slice, using its neighbors merely as contextual guides. Pioneering implementations of this technique are noted in the studies by~\cite{Roth2014new,Han2017}.

Furthermore, Generative Adversarial Networks (GANs) are employed in the segmentation of liver tissues. A 3D deep image-to-image network (DI2IN) is utilized for the liver task in~\cite{Yang2017}, while cascaded conditional GANs is the choice in~\cite{Chen2019}. In another example, ~\cite{Zheng2019} uses a GAN that integrates a deep atlas prior, where the generator  is based on DeepLab (ResNet101) for liver segmentation, and the discriminator is a straightforward 2D FCN tasked with evaluating the generator’s output. Similarly, tumor segmentation leveraging a combination of DeepLabV3 and Pix2Pix GAN is explored in~\cite{Xia2019}. Due to their complexity in training and fine-tuning, GAN-based approaches are less commonly pursued by researchers.

\subsubsection{U-Net-Based and U-Net-Like Models}

A significant advancement in FCN architecture within the biomedical sector is the U-Net model, pioneered by~\cite{Ronneberger2015}. This architecture features a symmetrical design with encoder and decoder layers, where the output of the encoder is merged into the decoder to enhance information flow during network training. This model has inspired a wave of innovative adaptations. For example, ~\cite{Chlebus2018} developed a variation called DenseU-Net for tumor segmentation, which incorporates a post-processing phase using random forests to decrease false positives. Additionally, ~\cite{Zhou2020} introduced U-Net++, a variant that incorporates an internal ensemble of DenseU-Nets to allow for layer-specific customizations.

Further advancements include the development of 3D versions of the U-Net to tackle the segmentation of liver tissues more effectively. ~\cite{Cicek2016} was among the first to expand U-Net into 3D, a concept further evolved by ~\cite{Zhang2018liver} and ~\cite{Li2018}, who implemented context-aware modules and hybrid 2D/3D models, respectively. Addressing computational efficiency, ~\cite{Zhang2018liver} integrated 3D patches into their U-Net variant tailored for multi-phase MRI analysis, while ~\cite{Mohagheghi2020} adapted a 3D U-Net to enhance the model through strategic modifications to its network structure.

In terms of application specificity, some studies like those by ~\cite{Zhang2018liver} and ~\cite{Li2018} have utilized a single model to segment both liver and tumors, illustrating the versatility of the U-Net architecture. Conversely, others have designed models tailored to specific tissues; for instance, ~\cite{wang2021sar} used a SAR-U-Net, which includes squeeze-and-excitation blocks and atrous spatial pyramid pooling, for precise liver segmentation from CT images. Additionally, specialized models for sequential tissue segmentation have been explored, as in ~\cite{Han2017}, where a DenseU-Net sequentially segments tumors following initial liver segmentation. This approach is also seen in the work of ~\cite{Jin2018} and ~\cite{Jiang2019}, where subsequent models build on the outputs of their predecessors to refine segmentation accuracy. Moreover,~\cite{Ouhmich2019} employs cascaded U-Net models to segment tumors following an initial liver segmentation, demonstrating a layered approach to tackling complex segmentation tasks.

In the study by Bai et al. \cite{Bai2019}, the liver segmentation produced by a 3D U-Net is enhanced through a multi-scale candidate generation (MCG) algorithm that identifies potential tumor regions using superpixels. These candidates are then processed using a 3D fractal residual network (FRN) coupled with an adaptive shape modeling (ASM) technique to refine the segmentation outcomes. Jansen et al. \cite{Jansen2019} employ a dual-pathway network to segment tumors from liver scans initially segmented by a 2D MC-FCN, utilizing nine-phase slices for this process. Zhang et al. \cite{Zhang2020a} initially segment the liver with a 2D U-Net and subsequently apply a 3D FCN for tumor segmentation, which is further refined using an LSM algorithm. In the work of Araujo et al. \cite{Araujo2021}, RetinaNet precedes a U-Net in generating an initial tumor segmentation, which is then refined by the U-Net and enhanced through various post-processing methods to improve the final segmentation accuracy.

Several innovative methods have emerged in recent studies, such as the work by~\cite{Tian2019} which introduces the global and local context U-Net (GLC-UNet). This model enhances the traditional U-Net by integrating both broader and more detailed contextual information, and it explores Couinaud segmentation of the liver. Similarly,~\cite{Fang2020deep} has developed a multi-input and multi-output feature abstraction network (MIMO-FAN), adapting the U-Net framework to handle inputs and outputs at multiple scales, thus preserving original information throughout the encoding processes.

In addressing the complexities of liver segmentation, modifications to the U-Net architecture have been specifically tailored for the unique structure of liver vessels. For instance,~\cite{montana2021vessel} applied a 2D U-Net incorporating batch normalization at each layer to segment liver vessels in 2D laparoscopic ultrasound (LUS) images.

The adoption of 3D networks has also been pivotal in enhancing the segmentation of anatomical structures.~\cite{Huang2018vessels} employed a 3D U-Net that is particularly effective in depicting narrow, tubular structures across slices. This study also tackled the issue of data imbalance with custom data augmentation strategies and specialized loss functions. Further,~\cite{li20213d} enhanced CT image segmentation of liver vessels using a novel 3D lightweight U-Net (LU-Net) integrated with a Graphical Connectivity Constraint Module (GCCM) that utilizes a graph attention network (GAT) to provide connectivity priors during training. Another groundbreaking approach is found in~\cite{hao2022hpm}, where a Hierarchical Progressive Multi-scale Network (HPM-Net) is introduced. This network leverages 3D U-Nets with a hierarchical multi-scale learning approach, combining internal and external progressive learning methods to better capture the varied semantic features of liver vessels. The design includes a dual-branch progressive (DBP) down-sampling technique to effectively enhance feature extraction capabilities.

Many liver vessel segmentation approaches utilize the U-Net architecture; however, not all features from the encoder are beneficial, leading to potential interference. To address this, Yan et al. \cite{yan2020attention} introduced LVSNet, a liver vessel segmentation network with specific designs that precisely depict the structure of vessels. A key feature of this network is the attention-guided concatenation (AGC) component, which adaptively selects and integrates relevant contextual features across different levels, focusing on capturing rich and detailed data effectively.

Other U-Net-like architectures have also been prominently featured in the literature. For example, Yang et al. \cite{yang2021liver} developed an inter-scale V-Net model designed to enhance liver vessel segmentation by employing dilated convolution to expand the receptive field without decreasing the rate of down-sampling and implementing a 3D deep supervision mechanism to accelerate convergence and enhance the learning of semantic features. This model incorporates inter-scale dense connections in the decoder to efficiently merge multi-scale features and preserve high-level semantic information.
Furthermore, Su et al. \cite{su2021dv} propose a dense V-Net (DV-Net) for liver vessel segmentation, which integrates dense blocks to improve feature learning and utilizes data augmentation from abdominal CT scans to augment the limited training data. The DV-Net also features a dual-branch dense connection down-sampling method and a combined Dice and BCE (D-BCE) loss function to optimize the capture of vascular features and maximize the utilization of imaging resources.

For a complete detailed review of the literature, the reader is advised to go through this review paper \cite{alkababji2023automated} published in 2023, where we summarize the literature, provide the lessons and gaps we uncovered, and our critical analysis of such methods.

\subsection{Medical Deployment Solutions}

\subsubsection{Standalone Application}
Researchers frequently integrate interactive features into their software, whether desktop or web-based, to allow user control over application operations or to enhance the performance of algorithmic segmentation. In~\cite{Arthurs2021}, the desktop application CardiovasculaR Integrated Modelling and SimulatiON (CRIMSON) is introduced. This tool provides a comprehensive, customizable, and user-friendly platform for conducting 3D and reduced-order computational hemodynamic studies. CRIMSON supports a range of functionalities: i) segmentation of vascular structures from medical images; ii) construction of analytic models for arterial and venous systems; iii) generation of finite element meshes; iv) design and implementation of boundary conditions; v) simulation of blood flow dynamics including fluid-structure interactions; and vi) post-processing and visualization of results such as velocity, pressure, and wall shear stress. Developed on the Medical Imaging Interaction Toolkit (MITK), CRIMSON was initially built using the Insight Toolkit (ITK), Visualization Toolkit (VTK), and Qt.

Another significant framework is the Interactive Learning and Segmentation Toolkit (ilastik), referenced in~\cite{Berg2019}. Ilastik enables domain experts to train classical machine learning-based biomedical image analysis tools through its Graphical User Interface (GUI). Once configured, the segmentation process in ilastik becomes fully automated, minimizing the need for further user interaction. This application is constructed using the PyQt, a Python-based Qt wrapper.

Moreover, ITK-SNAP, as detailed in~\cite{Yushkevich2019}, stands out as a seminal tool for interactive segmentation and visualization of medical images. Originally developed in the early 2000s, ITK-SNAP allows users to engage with anatomical structures in 3D images, supporting both manual and semi-automatic methods. Over time, it has been enhanced with additional features, some of which employ machine learning algorithms like random forests to assist in expert-guided annotations. Given the growing importance of machine learning in research, tools like ITK-SNAP that facilitate the creation of annotations are invaluable. ITK-SNAP is constructed using Qt and FLTK (pronounced "fulltick"), built on C++, and integrates medical processing libraries such as ITK and VTK.



\subsubsection{Add-on/Plugin}
Other scientists employ their applications as plugins or add-ons to already existing massive frameworks. That is the case for~\cite{Kordt2018}, where they utilize a GPU-based application for semi-automatic segmentation of brain tumors using a volumetric region-growing approach that the user can interactively tune. Their 3D region growing workflow is integrated within a 2D segmentation web application based on WebGL called VISIAN. Their work does not employ ML-based algorithms; however, it can play a significant role in creating manually-labeled brain tumor datasets as a reference for ML-based ones. In the study referenced by \cite{Arganda-Carreras2017}, the Trainable WEKA Segmentation (TWS) tool is presented, which originates from the Waikato Environment for Knowledge Analysis (WEKA). TWS is utilized as a pixel classification tool that offers a wide range of applications including boundary detection, object detection and localization, and semantic segmentation of microscopic objects. It takes advantage of a limited number of manually annotated images to automatically annotate the rest of the data using a library of methods supported by a GUI to ease the user experience (UX). It is also integrated into Fiji's segmentation plugins to analyze image data. Furthermore, researchers can quickly try out segmentation algorithms using TWS methods in Fiji with any available scripting languages.

Several studies incorporate standalone applications that utilize machine learning (ML), convolutional neural networks (ConvNets), and deep learning algorithms to address various challenges. Aleem et al.~\cite{Aleem2020} developed ML-based tools to track the impact of COVID-19 on axial CT scans of human lungs. They implemented the Mask RCNN framework with ResNet50 and ResNet101 as backbones to delineate areas affected by COVID-19, categorizing patient conditions as either "Mild" or "Alarming." The backend integrates Python, Flask, and PostgreSQL, while the frontend employs React and Material UI, all deployed on the Google Cloud Platform for enhanced scalability. Yang et al.~\cite{Yang2020} introduced NuSeT, a nuclear segmentation tool that applies deep learning to segment nuclei in microscopy images across various fluorescence imaging types. This tool combines a U-Net with a Region Proposal Network (RPN) and uses a watershed method for precise nuclear boundary detection in both 2D and 3D images. It features a straightforward yet effective graphical user interface (GUI) built on the TKinter Python library, allowing users to utilize or train TensorFlow (TF) modules directly from the GUI. Kim et al.~\cite{Kim2020} describe an automated web-based application that uses the U-Net for spine segmentation, trained on 330 CT images and evaluated on 14 images, using Keras to facilitate training. The GUI for this application is crafted using HTML, CSS, JavaScript, and the Bootstrap framework. Finally, Calisto et al.~\cite{Calisto2021} developed BreastScreening, a web application that leverages AI to assist in the breast cancer classification process. It supports multi-modality imaging and integrates a DenseNet ConvNet to provide automated and reliable classifications within the medical diagnostic workflow.
The front-end is built using HTML and jQuery (a JavaScript library), while it uses the PyTorch framework to build ConvNets and CornerstoneJS to execute functionalities asynchronously. The exciting part of their study is that they test AI's usefulness by allowing radiologists to use BreastScreening with and without AI integration. They conclude that radiologists are pleased with the AI's benefits, especially in reducing the cognitive workload and improving diagnosis execution. In~\cite{Song2020}, a GUI by the name FeAture Explorer (FAE) is created allowing researchers produce preprocessing steps and traditional supervised ML models that are developed to enhance their performance. The latter is evaluated via performance metrics embedded within the software. They develop candidate models to classify prostate cancer over the PROSTATEx dataset~\cite{Song2020} to show their case that the FAE application is practical. The software is built using PyQt as the basis for the GUI.

Recent investigations have explored the integration of virtual reality (VR) and augmented reality (AR) in the healthcare sector. Notably,~\cite{Chheang2021} discusses a collaborative VR platform developed to aid surgeons in planning liver tumor surgeries. This platform, crafted using the Unity engine, Virtual Reality Toolkit (VRTK), and Photon frameworks, enables both remote and co-located surgical teams to interact with patient-specific organ models. Users can manipulate these virtual models in both 3D and 2D representations to plan resections and assess potential risks. Additionally, the system provides a risk map visualization feature, which displays safety margins around tumors to guide hepatectomy procedures. In Trestioreanu~\cite{Trestioreanu2018}, the author opted for an AR framework with the help of Unity as the software and HoloLens as the hardware. In addition, he uses deep retinal image understanding (DRIU) ConvNet to segment both the liver and tumors over the liver tumor segmentation (LiTS) dataset. A ResNet50 plays the role of a detector ConvNet to reduce false positives before feeding the pixels into the DRIU for the tumor segmentation. The postprocessing involves using a 3D CRF to enhance the segmentation further. Consequently, the segmented liver and tumors can be portrayed by the AR headset the surgeon is wearing while operating on the patient.

The objectives differ significantly in the studies by~\cite{Chen2019deep,Yayuan2021} with respect to their intended audiences. Chen et al.~\cite{Chen2019deep} developed DeepLNAnno, short for \enquote{Deep Lung Nodules Annotation}, which is designed to identify and annotate lung nodules on CT images using a straightforward 3D ConvNet. This tool operates in a semi-automatic mode and was initially trained on datasets comprising adenocarcinoma and benign nodules. The annotations generated are particularly aimed at assisting Ph.D. holders who are not expert radiologists. Conversely, Yayuan et al.~\cite{Yayuan2021} introduced RadCloud, a platform that merges ML, deep learning, and radiomics analysis with robust data management capabilities. This tool is tailored for researchers and scientists who aim to develop sophisticated medical solutions. RadCloud utilizes cloud technology to provide a cost-effective, easy-to-use, and accessible data storage and sharing environment. It features user-friendly interfaces that facilitate automatic feature extraction, hyperparameter tuning, and the training, validation, and testing of ML and deep learning models.

Table~\ref{tab:LR_summary} encapsulates the findings from this literature review. A common thread across these studies is their focus on simplifying the use of statistical, ML, and deep learning models for non-programmers and clinicians in diverse applications. Our research similarly pursues this objective, specifically addressing the challenge of liver tissue segmentation. What sets our work apart is the creation of a comprehensive tool that sequentially segments all liver tissues, designed initially for the liver but adaptable to other organs. This tool segments the organ first, followed by tumors and vessels within it. Moreover, our application minimizes user input and assumes little prior technical knowledge, enhancing its accessibility and ease of use.

\begin{table*}[!ht]
    \caption{Desktop and web application related Work Summary.}
    \label{tab:LR_summary}
    \centering
    \resizebox{\linewidth}{!}{%
\begin{tabular}{
>{\centering\arraybackslash}m{0.13\textwidth} 
m{0.15\textwidth}
>{\centering\arraybackslash}m{0.1\textwidth}
m{0.18\textwidth}
>{\centering\arraybackslash}m{0.1\textwidth}
>{\centering\arraybackslash}m{0.06\textwidth}
>{\centering\arraybackslash}m{0.08\textwidth}
m{0.3\textwidth}
}
\toprule
Ref.                        & \multicolumn{1}{c}{Abbrv.} & Platform           & \multicolumn{1}{c}{GUI Framework}   & ML usage & Organ                  & Auto       & \multicolumn{1}{c}{Functionality}                                                     \\
\midrule
\cite{Arthurs2021}          & CRIMSON                    & Desktop            & MITK + PyQt                         & No       & Vessels                & No         & Generates 3D haemodynamics studies                                                 \\
\cite{Berg2019}             & ilastik                    & Desktop            & PyQt                                & Yes      & Cell Nuclei            & Yes        & Interactive software to segment, classify, track, and count cells                     \\
\cite{Yushkevich2019}       & ITK-SNAP                   & Desktop            & Qt + FLTK                           & Yes      & Generic                & Semi       & New features addition to ITK-SNAP tool   \\
\cite{Kordt2018}            & N/A                        & Web                & VISIAN                              & No       & Brain                  & Semi       & Interactive segmentation for brain tumor              \\
\cite{Arganda-Carreras2017} & TWS                        & Desktop            & Plugin in Fiji                      & Yes      & Cells                  & Yes        & A plugin in Fiji framework to add WEKA initiative    \\
\cite{Aleem2020}            & N/A                        & Web                & React + Material UI                 & Yes      & Lungs                  & Yes        & COVID-19 case testing over lungs                                                      \\
\cite{Yang2020}             & NuSeT                      & Desktop            & Tkinter                             & Yes      & Cell Nuclei            & Yes        & Automated training on and segmentation of nuclei cells using AI                          \\
\cite{Kim2020}              & N/A                        & Web                & HTML + CSS + JavaScript + Bootstrap & Yes      & Spine                  & Yes        & Spine segmentation web application using U-Net ConvNet                                \\
\cite{Calisto2021}          & Breast\-Screening            & Web                & HTML + jQuery (JavaScript)          & Yes      & Breast                 & Yes        & AI utilization in breast images cancer classification                                 \\
\cite{Song2020}             & FAE                        & Desktop            & PyQt                                & Yes      & Pancreas               & Yes        & Prostate cancer classical ML classifier with engineered features     \\
\cite{Chheang2021}          & N/A                        & Desktop Web Server & Unity + VRTK + Photon               & No       & Liver                  & No         & VR for liver's tumors resection                                                       \\
\cite{Trestioreanu2018}     & N/A                        & Desktop            & Unity + HoloLens                    & Yes      & Liver                  & Yes        & Liver \& tumor segmentation with AR capabilities                                     \\
\cite{Chen2019deep}         & DeepLNAnno                 & Web                & Not Disclosed                       & Yes      & Lungs                  & Semi       & Lung nodules annotating system over CT-based images                                   \\
\cite{Yayuan2021}           & RadCloud                   & Web/\-Mobile         & Not Disclosed                       & Yes      & Generic                & Yes        & Facilitate access for AI-assisted research for different challenges \\
\bottomrule
\end{tabular}}
\end{table*}

\section{Methodology}\label{sec3}
\begin{figure}[!ht]
    \centering
    \includegraphics[width=\linewidth, trim={1.3in 1.7in 1.3in 1.7in}, clip]{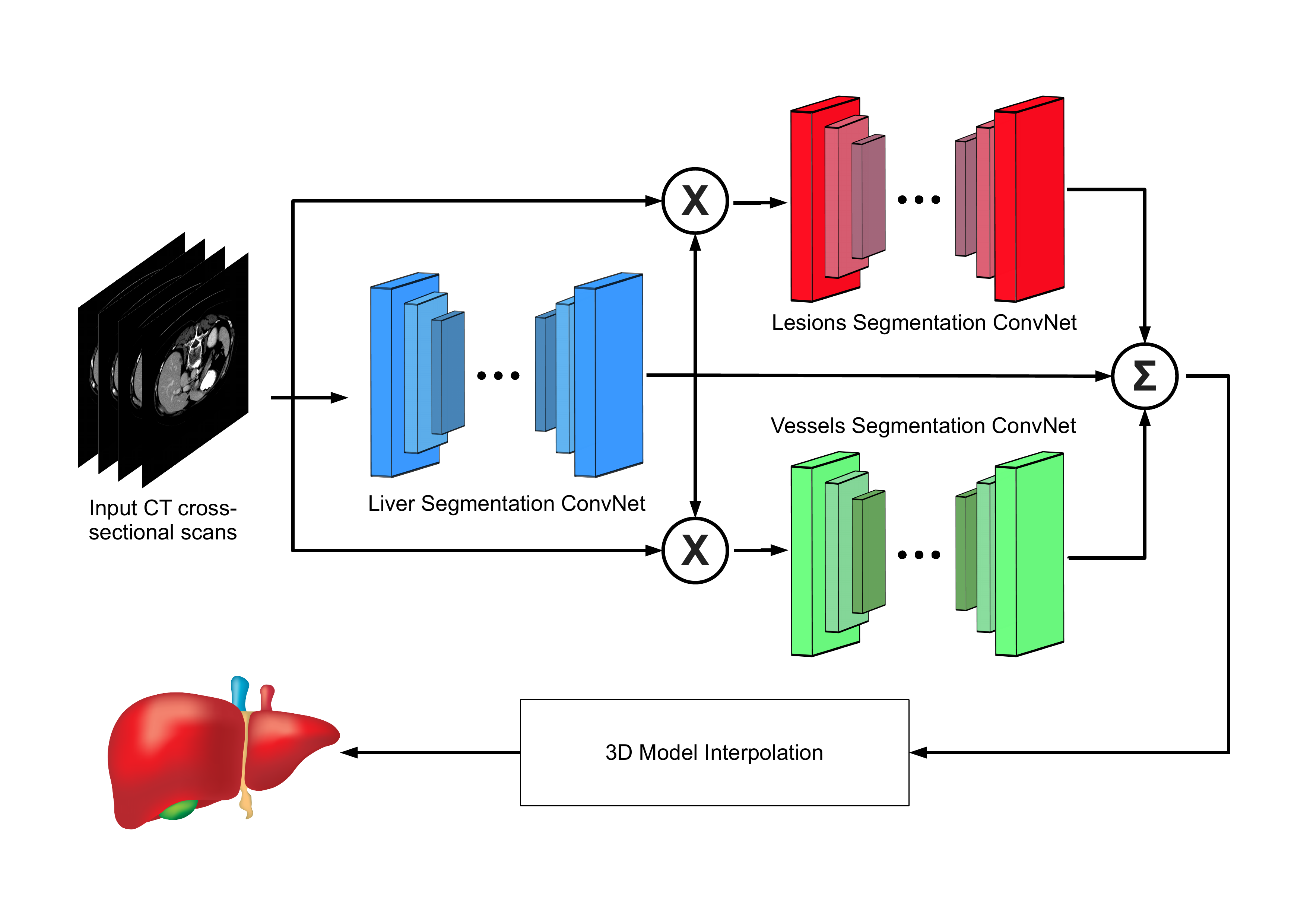}
    \caption{Overall system design where liver ConvNet, tumors ConvNet, and vessels ConvNet are depicted by blue, red, and green, respectively.}
    \label{fig:overall-system}
\end{figure}

This section outlines the datasets employed, image preprocessing methods, the neural network architecture, the training setup, evaluation criteria, and the processes for 3D modeling and printing. The system addresses liver tissue segmentation by initially processing the CT scans through a ConvNet dedicated to liver segmentation. The resulting masks are applied to the original CT scans to isolate the liver voxels and eliminate extraneous tissue. Subsequently, these refined slices are input into separate ConvNets designed for tumor and vessel detection, which operate concurrently to produce specific tissue masks. Finally, the combined outputs from all ConvNets are merged and transformed into a 3D model of the liver, ready for visualization or printing. Figure~\ref{fig:overall-system} succinctly illustrates this workflow.

The logic behind opting for such methodology is to allow the tumors and vessels ConvNets to focus on a region of voxels contained within the liver, avoiding training for over-generalizability that simply is not useful, given the fact a liver-segmenting ConvNet.will always be present.

\subsection{Utilized Dataset and ConvNet Architecture}
The focus of this paper is on segmenting the liver parenchyma, tumors, and vessels, necessitating a dataset with accurate ground-truth segmentation for these tissues. According to~\cite{alkababji2023automated}, the ideal dataset for this purpose is identified as the Medical Segmentation Decathlon Challenge: Task 8 Hepatic Vessel (MSDC-T8). This dataset includes 443 contrast-enhanced (CE) CT scans, with physical spacings ranging from 0.8 to 8mm between slices and 0.58 to 0.97 mm within slices. The resolution between slices influences the total number of slices generated—higher spacings result in fewer slices, and vice versa, with totals varying between 24 and 251 slices per volume. The liver ground-truth labels were contributed by Tian et al.~\cite{Tian2019}, while the ground-truth masks for tumors and vessels were provided by the challenge organizers in~\cite{simpson2019large}. While liver masks are available for all 443 records, tumor and vessel masks are available for 303 of these.

\begin{figure}[t!]
    \centering
    \includegraphics[width=\textwidth, trim={1.5in 0in 0.8in 0in}, clip]{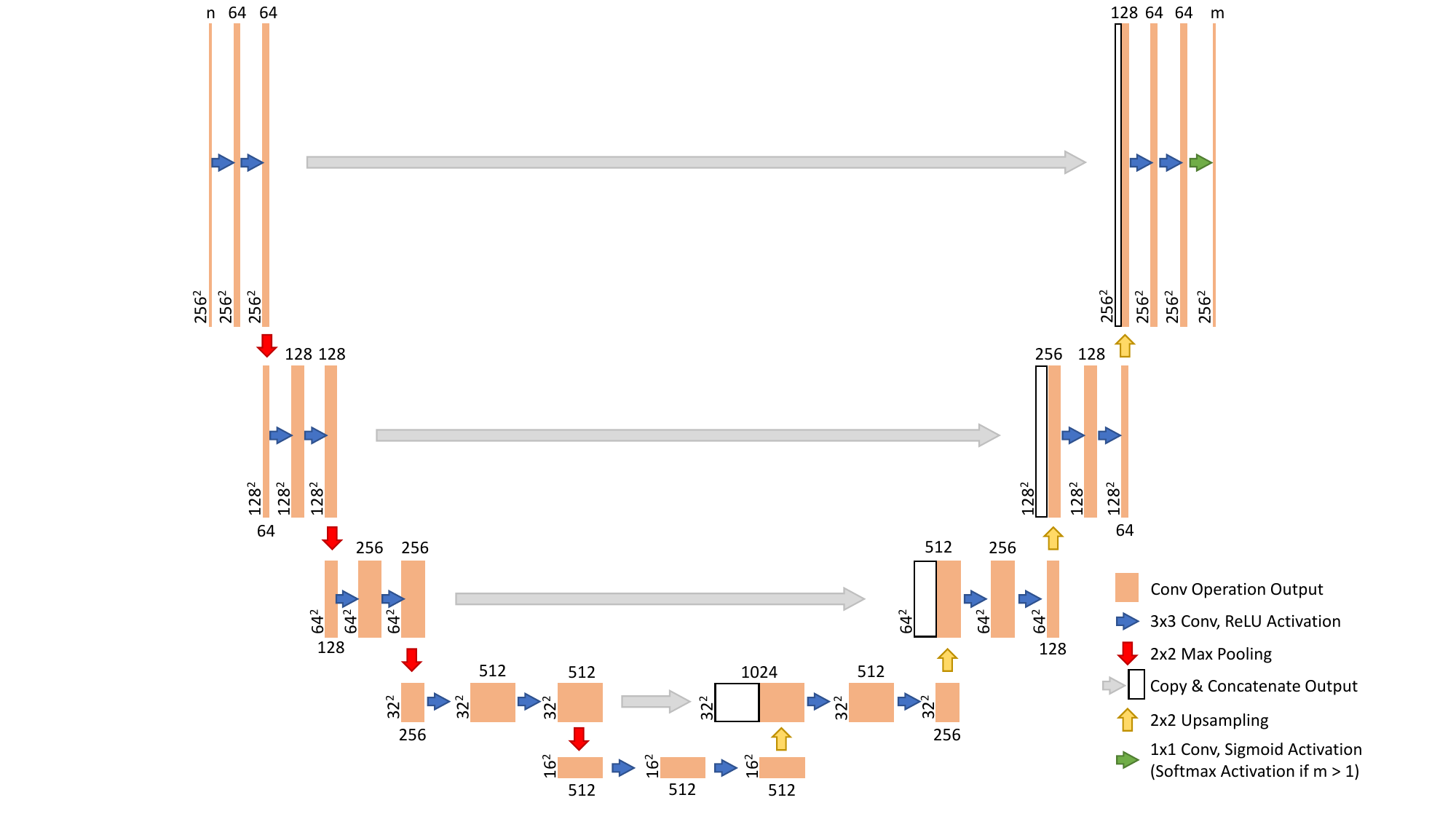}
    \caption{Chosen U-Net ConvNet with slight modifications.}
    \label{fig:U-net}
\end{figure}

For the neural network, a U-Net architecture is employed, optimized for this application as depicted in Figure~\ref{fig:U-net}. The chosen configuration uses standard filter size $f$~=~3 (i.e., $3\times3$), with padding $p$~=~1, and stride $s$~=~1 to preserve the dimensions of the convoluted layers, allowing for straightforward integration of encoder outputs with corresponding decoder components without the need for cropping. A key modification in our implementation is the number of input channels; our U-Net processes $n$ slices (where $n=2k+1$ for 2.5D input), diverging from the original single-slice input format. It is crucial to note that the complexity and the number of parameters can significantly increase with deeper networks. For instance, we reduced the number of filters at the deepest layer from 1024 to 512, cutting the parameters required for each convolution at this level from 9,438,208 to 2,359,808, thus making the model more manageable and less resource-intensive.

It has been adapted from the source, where we modify the input slices' dimensions and the number of channels that the input layer accepts. Unlike the original U-Net, we modify the U-Net's last level to reduce the number of parameters.

\subsection{Data Preprocessing}
\begin{figure}
    \centering
    \includegraphics[width=\textwidth]{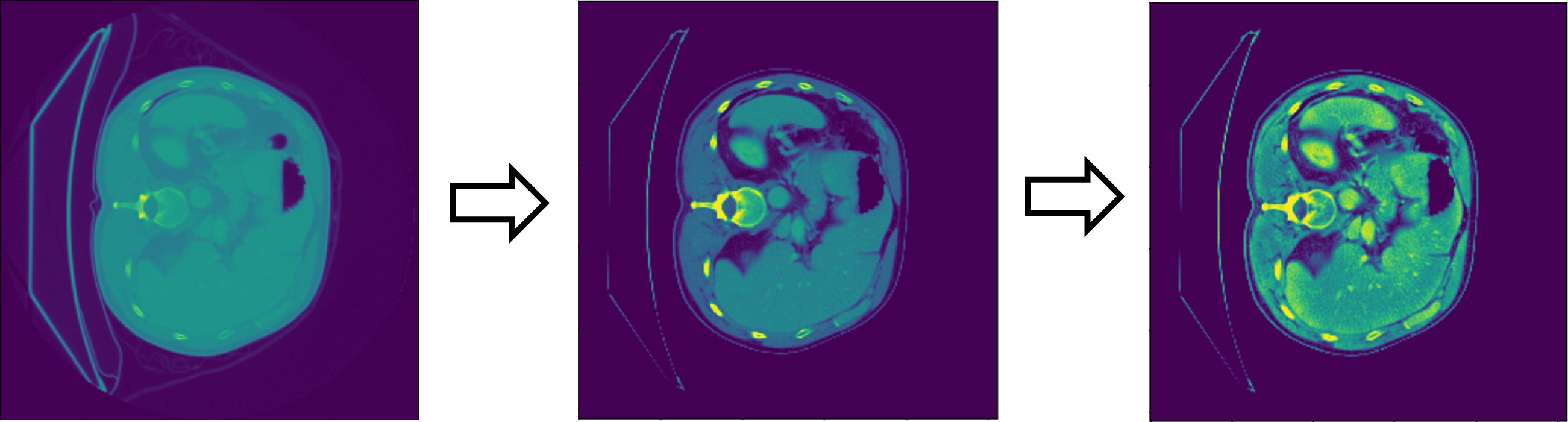}
    \begin{subfigure}[b]{0.25\textwidth}
    \caption{}
    \label{fig:pre_processing_demo_a}
    \end{subfigure}
    \hfill
    \begin{subfigure}[b]{0.35\textwidth}
    \caption{}
    \label{fig:pre_processing_demo_b}
    \end{subfigure}
    \hfill
    \begin{subfigure}[b]{0.25\textwidth}
    \caption{}
    \label{fig:pre_processing_demo_c}
    \end{subfigure}
    \caption{Preprocessing highlights a) original slice, b) after applying intensity clipping, and c) after applying CLAHE.}
    \label{fig:pre_processing_demo}
\end{figure}

\begin{figure}[!ht]
    \centering
    \begin{subfigure}[b]{0.49\textwidth}
    \includegraphics[width=\linewidth]{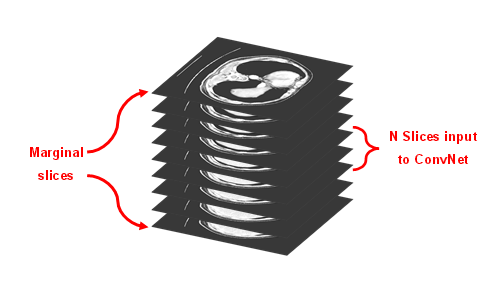}
    \caption{2.5D input concept.}
    \label{fig:preprocessed-data-results-a}
    \end{subfigure}
    \hfill
    \begin{subfigure}[b]{0.5\textwidth}
    \includegraphics[width=\linewidth]{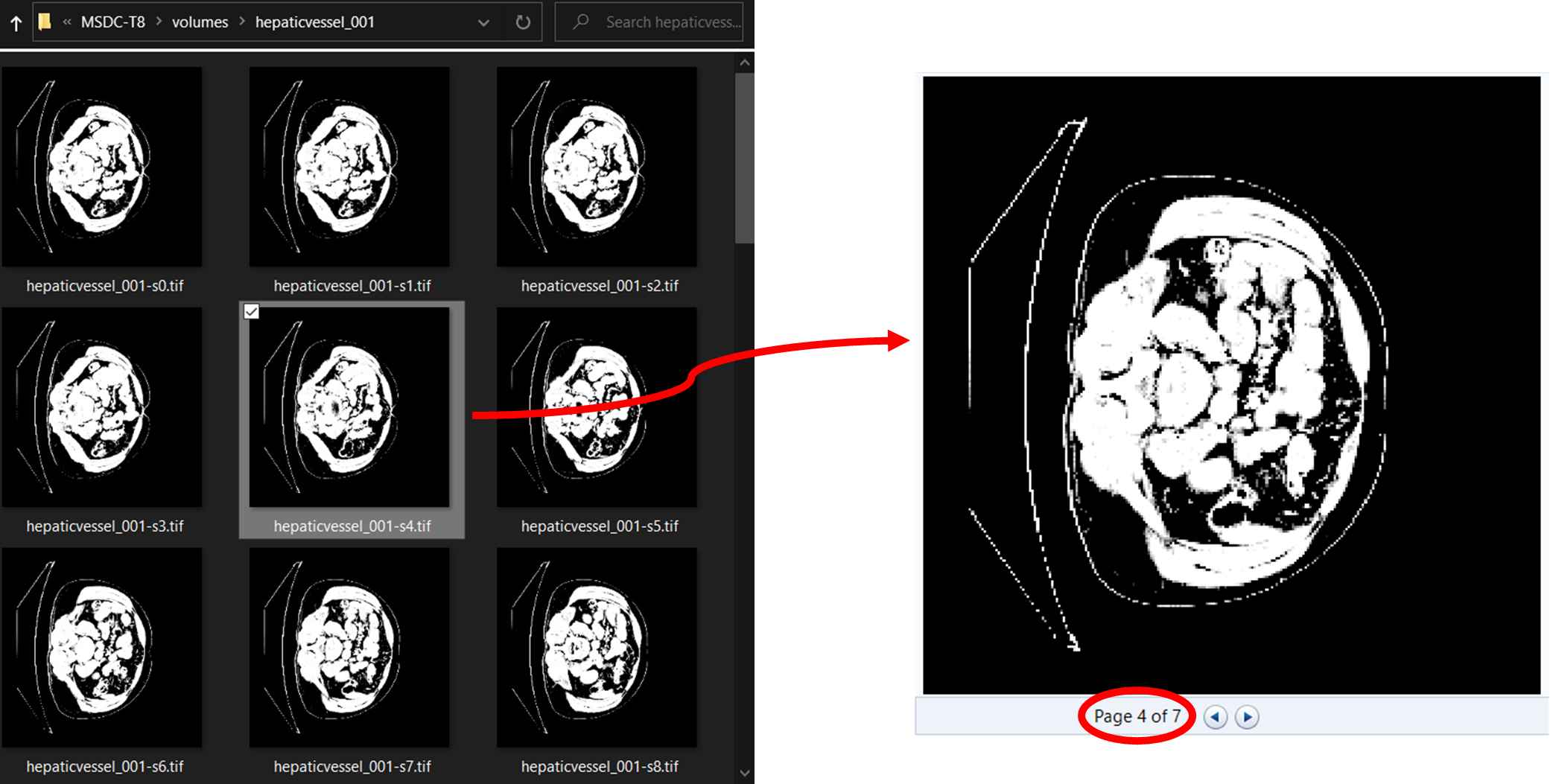}
    \caption{Resulting preprocessed 2.5D dataset.}
    \label{fig:preprocessed-data-results-b}
    \end{subfigure}
    \caption{Preprocessing highlights a) original slice, b) after applying intensity clipping, and c) after applying CLAHE.}
    \label{fig:preprocessed-data-results}
\end{figure}

The nibabel Python library is utilized to read the data from .nii.gz files. The following set of preprocessing techniques are applied to the CT scans in the following order before feeding them into the ConvNets to boost their performance:

\begin{enumerate}
\item \textbf{Volumetric Rotation:} Due to the diverse scanning protocols across various medical facilities, CT scans may be oriented differently. The orientation data is accessible within the metadata of the NIfTI files. To standardize this and reduce unnecessary variability, CT scans and their corresponding masks are adjusted to a uniform orientation (to the right). This standardization helps in focusing on the essential variability rather than accommodating all possible orientations of patient scans.

\item \textbf{Volumetric Rescaling:} To manage a large number of parameters in ConvNets, such as the U-Net, and to streamline the computational demands of the network, a rescaling factor of $\dfrac{1}{2}$ is applied to each dimension of the input CT scans. Without rescaling, each input would have a dimension of 512×512 pixels (263,680 total pixels), but with rescaling, dimensions are reduced to 256×256 pixels (65,536 total pixels), effectively reducing the computational load by approximately $\dfrac{1}{4}$. This rescaling is implemented across all training, validation, and testing images to minimize computational complexity without altering the number of parameters that need fine-tuning. During testing, however, the downsampling step is omitted for the masks to ensure that the segmented outputs can be compared accurately to their original dimensions (512×512 pixels), closely mirroring the clinical settings. This approach also enhances the resolution of the 3D models due to the increased voxel count.

\item \textbf{Voxels' Intensity Clipping:} As identified by Yuan~\cite{Yuan2017}, liver voxel intensities typically range from -100 to 400. By clipping the voxel intensity values outside these bounds, the contrast between the liver and other tissues, as well as among different organs, is enhanced. The equation in~\eqref{eq:clipping} specifies the technique used for clipping. The beneficial impact of this preprocessing step is illustrated in Figures~\ref{fig:pre_processing_demo_a} and \ref{fig:pre_processing_demo_b}.

    \begin{equation} \label{eq:clipping}
    V_{(x,y,z)} = \begin{cases} 
      \multicolumn{1}{@{}c@{\quad}}{-100} & \multicolumn{1}{@{}c@{\quad}}{V_{(x,y,z)} \leq -100} \\
      \multicolumn{1}{@{}c@{\quad}}{V_{(x,y,z)}} & \multicolumn{1}{@{}c@{\quad}}{-100 < V_{(x,y,z)} < 400} \\
      \multicolumn{1}{@{}c@{\quad}}{400} & \multicolumn{1}{@{}c@{\quad}}{400 \leq V_{(x,y,z)}}
      \end{cases}
    \end{equation}
\noindent where $V_{(x,y,z)}$ denotes the intensity of a voxel in volume $V$ at coordinates $(x, y, z)$.

\item \textbf{Voxels' Intensity Range Standardization:} Diverse datasets like MSDC-T8 often include volumes with disparate intensity ranges for voxels, such as from -1100 to 3024 or -1000 to 1200. These discrepancies often arise from the use of different imaging equipment, each adhering to distinct calibration standards. To unify these, all volumes undergo a standardization process that normalizes their values to a consistent range from 0 to 1, according to the formula given in equation~\eqref{eq:standardization}:
\begin{equation} \label{eq:standardization}
V_{(x,y,z)} = \dfrac{V_{(x,y,z)} - \min(V_{(x,y,z)})}{\max(V_{(x,y,z)}) - \min(V_{(x,y,z)})}
\end{equation}
Considering the earlier application of clipping, which restricts voxel values to the [-100, 400] interval, the normalization process simplifies to a subtraction of -100 followed by division by 500.

\item \textbf{Contrast Enhancement via CLAHE:} To better delineate the boundaries between different organs and internal liver structures such as parenchyma, tumors, and blood vessels, Contrast-Limited Adaptive Histogram Equalization (CLAHE) is employed. Unlike traditional histogram equalization that might not perform well across areas with high contrast variation, CLAHE adjusts histograms locally, using small regions around each pixel/voxel. This method is particularly effective in balancing the brightness within images that contain areas significantly darker or lighter than the average. The scikit-image Python library provides support for applying CLAHE on 3D data arrays, which is advantageous for analyzing medical images that include sequential slices. The impact of CLAHE application is demonstrated in Figures~\ref{fig:pre_processing_demo_b} and \ref{fig:pre_processing_demo_c}, particularly visible on slice 25 of the volume labeled \enquote{hepaticvessel\_001}.

\item \textbf{Voxels' Normalization:} For faster training convergence, normalization is applied over all the records' voxels. It is implemented by applying equation~\eqref{eq:normalization} over all the voxels available within the dataset.
\begin{equation} \label{eq:normalization}
V_{(x,y,z)} = \dfrac{V_{(x,y,z)} - \mu_V}{\sigma_V}
\end{equation}
\noindent where $\mu_V$ and $\sigma_V$ represent the mean and standard deviation, respectively, of voxel intensities across all examined volumes.

\item \textbf{Volume Slicing:} In preparation for ConvNet processing, the original 3D data sets are converted into 2D or 2.5D formats. This is achieved by converting the NIfTI volumes into Tag Image File Format (TIFF) files. TIFF is chosen for its ability to maintain the integrity of floating point data without information loss. Additionally, TIFF supports the storage of multiple images within a single file, which facilitates the creation of localized 3D volume segments. This capability is particularly valuable for generating 2.5D inputs tailored for ConvNet applications. In the case of inputting 2.5D data, the slice we aim to segment is the middle slice, where the adjacent slices serve as volumetric context providers for the ConvNet. It essentially groups adjacent slices in groups of $2k+1$ ($k \in \mathbb{Z}$) as shown in Figure~\ref{fig:preprocessed-data-results}. The resultant of the aforementioned data preprocessing steps. To create a 2.5D image for the marginal slices shown in Figure~\ref{fig:preprocessed-data-results-a}, the marginal slice itself gets repeated $k$ times. The close-to-margin slices get treated in the same way to have a consistent $2k+1$ 2.5D input shape.
\end{enumerate}

\subsection{3D Construction and Printing}
After developing three Convolutional Neural Networks (ConvNets) specialized in segmenting liver parenchyma, tumors, and vessels, the generation of a 3D model representing these essential liver tissues has become feasible. Utilizing the "Marching Cubes" algorithm (referenced from scikit-image), we interpolate the 2D cross-sectional slices into their corresponding 3D forms. This approach is deemed appropriate for our focus on implementing the AI Radiologist system within CT scans. However, to consolidate the objects into a unified .obj and .mtl file, we undertake the task entirely in Python.

\begin{figure}
    \centering
    \includegraphics[width=0.75\textwidth]{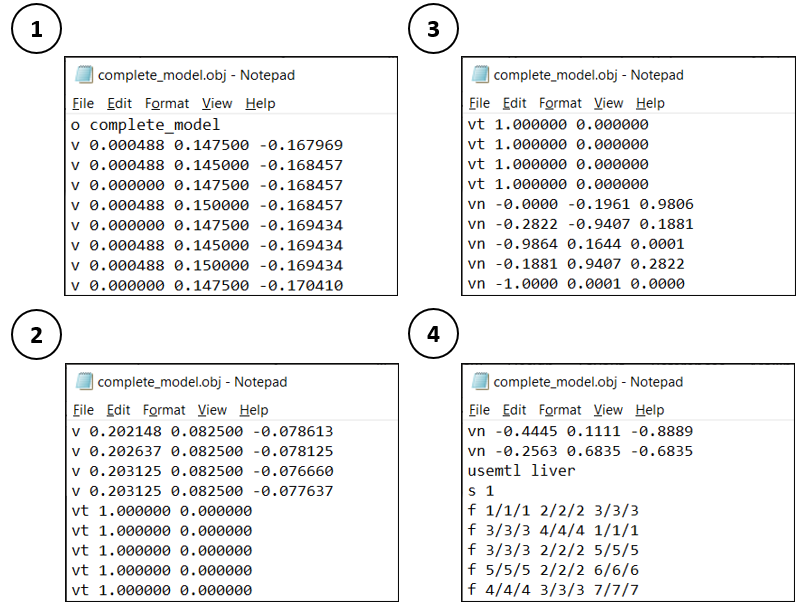}
    \caption{Mandatory components of a .obj file.}
    \label{fig:obj_file_structure}
\end{figure}

Initially, the masks of the three tissues are imported into the script following their generation by the designated ConvNets, facilitated by the nibabel Python library. Subsequently, the 3D Lewiner interpolation, employing the "Marching Cubes" algorithm, is utilized to construct the 3D object representation of each tissue, resulting in the creation of the .obj file. It's noteworthy that the .obj file specifies four essential parameter types: vertex (v), vertex texture (vt), vertex normal (vn), and face (f). A detailed breakdown of these parameters required to construct the .obj file is illustrated in Figure~\ref{fig:obj_file_structure}.

\begin{figure}[!tp]
    \centering
    \includegraphics[width=0.7\textwidth]{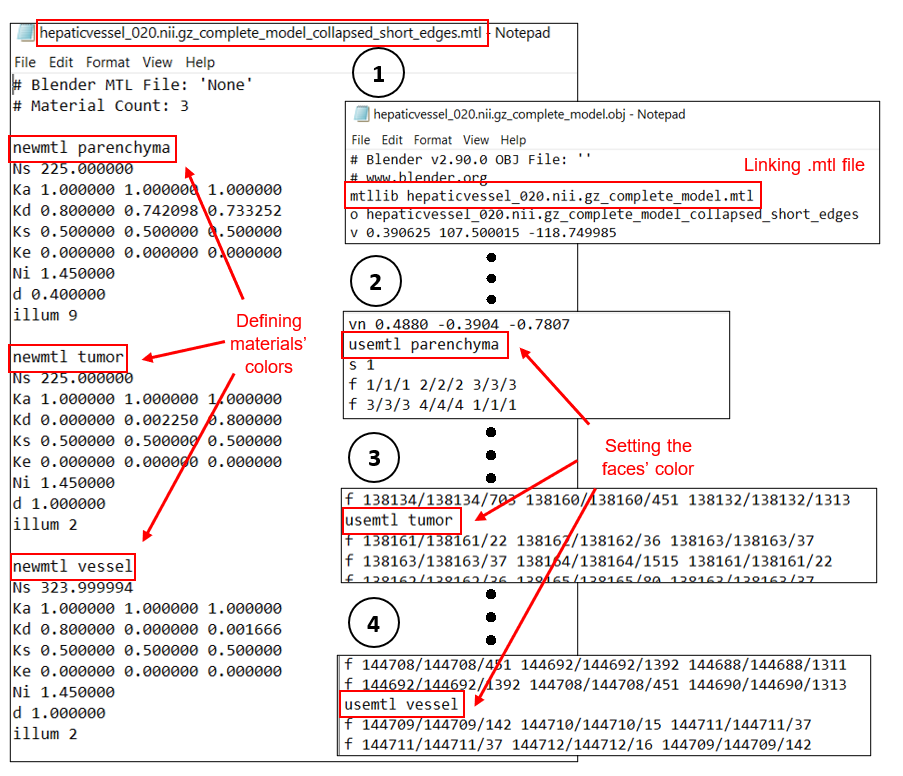}
    \caption{Resulting .obj and .mtl files main components.}
    \label{fig:mtl_inclusion}
\end{figure}

The coordinates of vertices in the x, y, and z 3D space are defined by the (\textit{v}s). Conversely, the importance of (\textit{vt}s) and (\textit{vn}s) is relatively lower. However, (\textit{vt}s) contribute to enhancing the realism of the 3D object by applying a pre-defined texture map to the faces, while (\textit{vn}s) assist in specifying the direction of the normal vector \cite{Chakravorty}. The (\textit{f}s) combine these components in a prescribed order, as depicted by equation~\eqref{eq:obj_face}:

\begin{equation} 
    f \quad v_1 [/vt_1][/vn_1] \quad v_2 [/vt_2][/vn_2]\quad v_3 [/vt_3][/vn_3] \quad \cdot\cdot\cdot \quad v_x [/vt_x][/vn_x] 
\label{eq:obj_face}
\end{equation}
In this context, $x$ signifies the variable number of vertices assigned to each face $f$, as indicated by the optional nature of these values in equation~\eqref{eq:obj_face}. Given that the scikit-image library is configured to produce vertices specifically for triangular faces, a trio of vertices is necessary for each face. During the creation of the .obj file for different tissues, the line \enquote{mtllib $<$material\_filename$>$.mtl} is added to establish a connection with the corresponding .mtl file. This connection is crucial for applying predefined textures and colors to various surfaces. Moreover, the directive \enquote{usemtl $<$tissue\_material$>$} is placed before the object's faces within the file, enabling the application of specific material attributes to the designated object.
As part of compiling .obj files for various tissues into a unified file, all vertex coordinates (\textit{v}s) are listed first, followed sequentially by texture coordinates (\textit{vt}s) and then normal vectors (\textit{vn}s). Faces (\textit{f}s) are added last. With the inclusion of additional (\textit{v}s), (\textit{vt}s), and (\textit{vn}s) within the same file, it becomes necessary to adjust the indexing for faces (\textit{f}s) to account for vertices that pertain to previously listed objects. For instance, as demonstrated in Figure~\ref{fig:mtl_inclusion}, the tumor object begins utilizing vertices at the 1380161\textsuperscript{st} position, following the 1380160 vertices allocated for the liver structure.

\subsection{AI Radiologist Desktop Application}


In this subsection, we explore the graphical user interface (GUI) design of the preliminary schematic for the desktop application and the underlying framework used for its implementation. The application is engineered with a focus on user-friendliness, intuitiveness, and delivering a high-quality user experience (UX).

\begin{figure}[!hbtp]
    \centering
    \includegraphics[width=0.6\textwidth, trim={0.3in 0.1in 0.3in 0.2in}, clip]{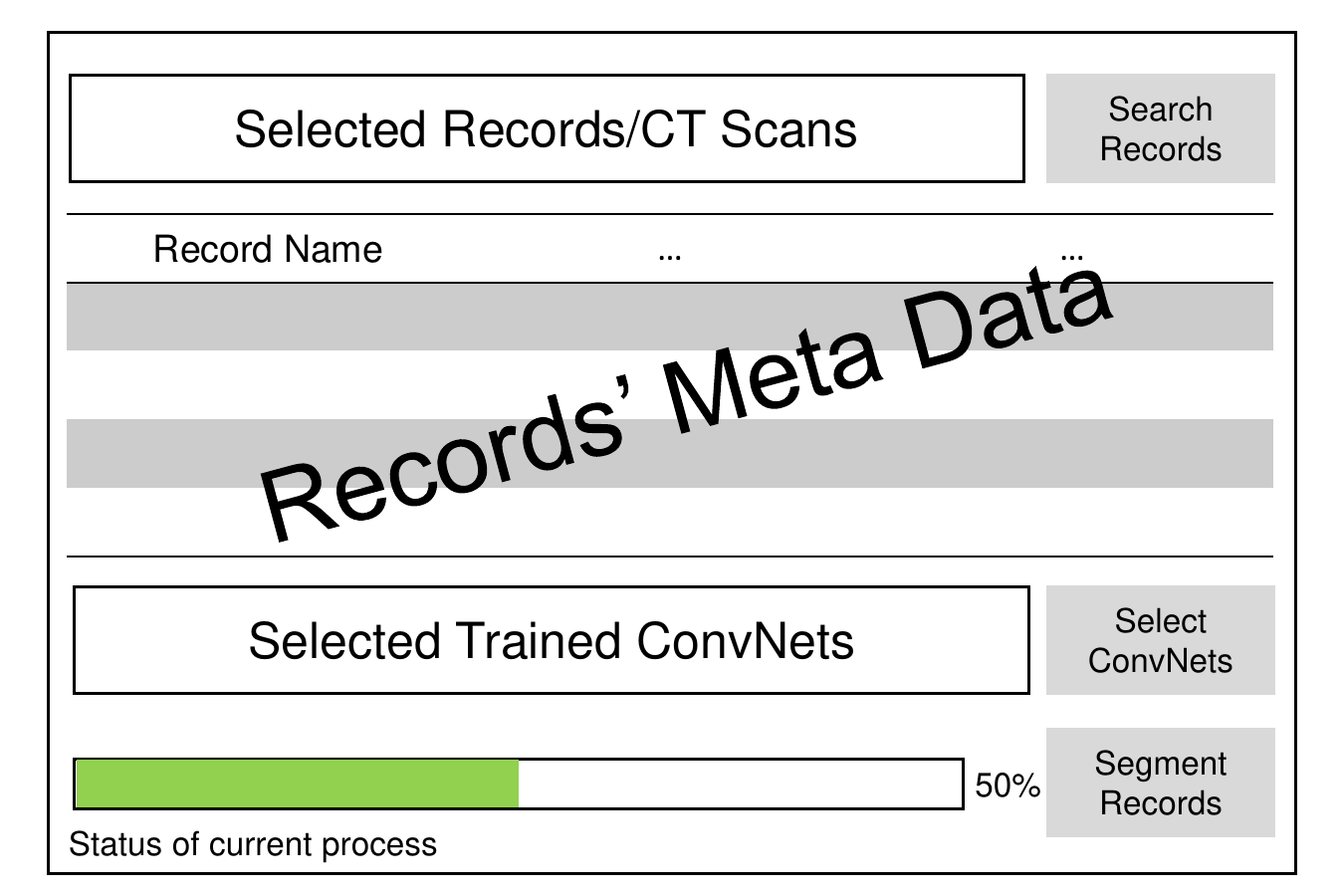}
    \caption{Preliminary design of the AI Radiologist desktop application.}
    \label{fig:application_schematic}
\end{figure}

Figure~\ref{fig:application_schematic} illustrates the initial design of the AI Radiologist application. To initiate segmentation, users are required to select the desired records by activating the \enquote{Search Records} button. This action triggers a pop-up window where users can sequentially choose multiple records for segmentation. Once records are selected, a table displaying CT record metadata is presented, enabling clinicians to review pertinent information such as voxel dimensions and slice counts. Furthermore, customization of the displayed metadata is possible to accommodate specific clinician preferences.

Subsequently, the Convolutional Neural Networks (ConvNets) trained for segmenting liver tissues can be designated by clicking the \enquote{Select ConvNets} button, which prompts another window. Upon selection, users can proceed to segment tissues sequentially using the \enquote{Segment Records} option. Progress is visually indicated through a progress bar integrated into the application interface, accompanied by status text messages, enhancing the overall user experience.

The application is intended to work offline as this approach requires less development. Arguably, it could be considered much safer than online platforms since it could work on any offline device that no hacker can reach.

\section{Results and Discussion} \label{sec4}

\subsection{Optimal Input Shape}


A review of existing research indicates that various input shapes have been incorporated into different ConvNets' architectures. This section explores the most effective input shape using a 2D ConvNet framework. The focus here is on 2D and 2.5D configurations, as these are straightforward to generate and integrate into the ConvNet. The comparative analysis of 2D (1 slice) and 2.5D (3, 5, and 7 slices) input shapes is detailed in Table~\ref{tab:optimal-input-shape}.

The analysis reveals that the 2.5D configuration with 5 slices represents the most efficient input shape for the liver segmentation task on the MSDC-T8 dataset. Performance enhancements were observed up to the 5-slice configuration, after which a slight reduction in effectiveness was noted with the inclusion of 7 slices. The ConvNet shows satisfactory performance across various input configurations; however, the 2.5D (5 slices) setup emerges as the ideal. This setup provides sufficient volumetric context while avoiding the overload of excessive global information necessary for segmenting more localized data, particularly where the spacing between slices is large enough to influence volumetric context understanding. Metrics such as Average Surface Distance (ASD), Root Mean Square Distance (RMSD), and Hausdorff Distance (HD) indicate higher error rates in the 2D configurations, underscoring the benefits of the 2.5D input shape in capturing volumetric context more effectively.

\begin{table}[!ht]
    \centering
    \caption{Optimal input shape investigation for the U-Net}
    \begin{tabular}{
m{0.15\textwidth}
>{\centering\arraybackslash}m{0.08\textwidth}
>{\centering\arraybackslash}m{0.08\textwidth}
>{\centering\arraybackslash}m{0.08\textwidth}
>{\centering\arraybackslash}m{0.08\textwidth}
>{\centering\arraybackslash}m{0.08\textwidth}
>{\centering\arraybackslash}m{0.08\textwidth}
>{\centering\arraybackslash}m{0.08\textwidth}
>{\centering\arraybackslash}m{0.08\textwidth}
}
\hline
\textbf{Input shape} & \textbf{Dice (\%)} & \textbf{IoU (\%)} & \textbf{RVD} & \textbf{ASD (mm)} & \textbf{RMSD (mm)} & \textbf{HD (mm)} & \textbf{95\% HD (mm)} & \textbf{Epoch} \\ \hline
\textbf{2D (1 slice)}  & 97.84 (0.21)          & 95.81 (0.38)          & -0.01 (0.003)           & 0.902 (0.564)          & 3.16 (1.62)          & 35.94 (5.65)          & 6.49 (5.35)          & \textbf{18.6 (3)} \\
\textbf{2.5D (3 slices)} & 98.04 (0.07)          & 96.17 (0.13)          & -0.01 (0.003)           & 0.596 (0.184)          & 2.34 (0.81)          & \textbf{26.94 (3.48)} & \textbf{2.66 (0.46)} & 22 (1.7) \\
\textbf{2.5D (5 slices)} & \textbf{98.12 (0.04)} & \textbf{96.33 (0.07)} & \textbf{-0.008 (0.002)} & \textbf{0.624 (0.443)} & \textbf{2.15 (1.40)} & 27.16 (4.53)          & 4.10 (4.37)          & 22.0 (1.6) \\        
\textbf{2.5D (7 slices)} & 98.02 (0.13)          & 96.15 (0.24)          & -0.008 (0.005)          & 0.763 (0.639)          & 2.41 (1.5)           & 29.73 (1.71)          & 4.48 (4.38)          & 20.2 (2.5)  \\ \hline  
\end{tabular}
    \label{tab:optimal-input-shape}
\end{table}

\subsection{Liver Segmentation Challenge Scheduling Techniques and Learning Rate (LR) Variations}
Learning rate (LR) schedulers are verified techniques that enhance the resulting model trained on a dataset. The idea behind those schedulers is that they manipulate the LR during the training journey, which allows it to better converge to a local minimum for a specific objective/loss function. 
In this section, we examine various effective learning rate (LR) schedulers while maintaining other variables constant to ensure a fair comparison. The focus here is on contrasting two renowned schedulers: ReduceLRonPlateau \cite{PyTorch-reducelronplateau} and OneCycleLR \cite{PyTorch-onecyclelr}, each built on a unique foundational philosophy. ReduceLRonPlateau adjusts the LR downwards as the model nears a local minimum in the cost function, optimizing model performance and fine-tuning the weights to closely match this local minimum. In contrast, OneCycleLR is designed to prevent the model from being ensnared by suboptimal local minima due to overfitting. It achieves this by strategically increasing the LR temporarily, steering the model towards a more advantageous minimum of the cost function. This approach is often considered a regulatory step in model training \cite{Smith2019}.

\begin{figure}[!ht]
    \centering
    \begin{subfigure}[b]{0.49\textwidth}
    \includegraphics[width=\linewidth]{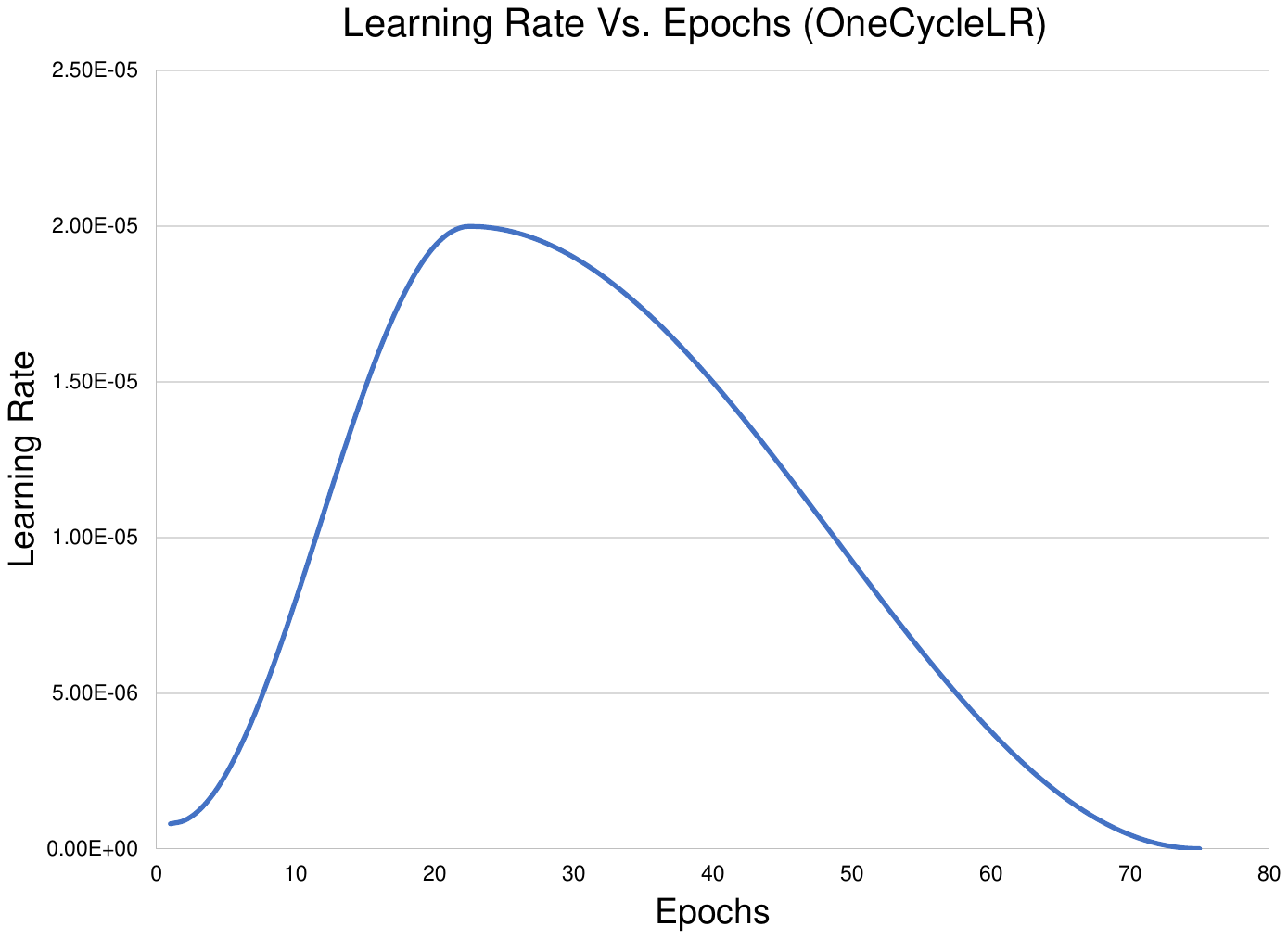}
    \caption{}
    \label{fig:schedulers-behavior-a}
    \end{subfigure}
    \hfill
    \begin{subfigure}[b]{0.5\textwidth}
    \includegraphics[width=\linewidth]{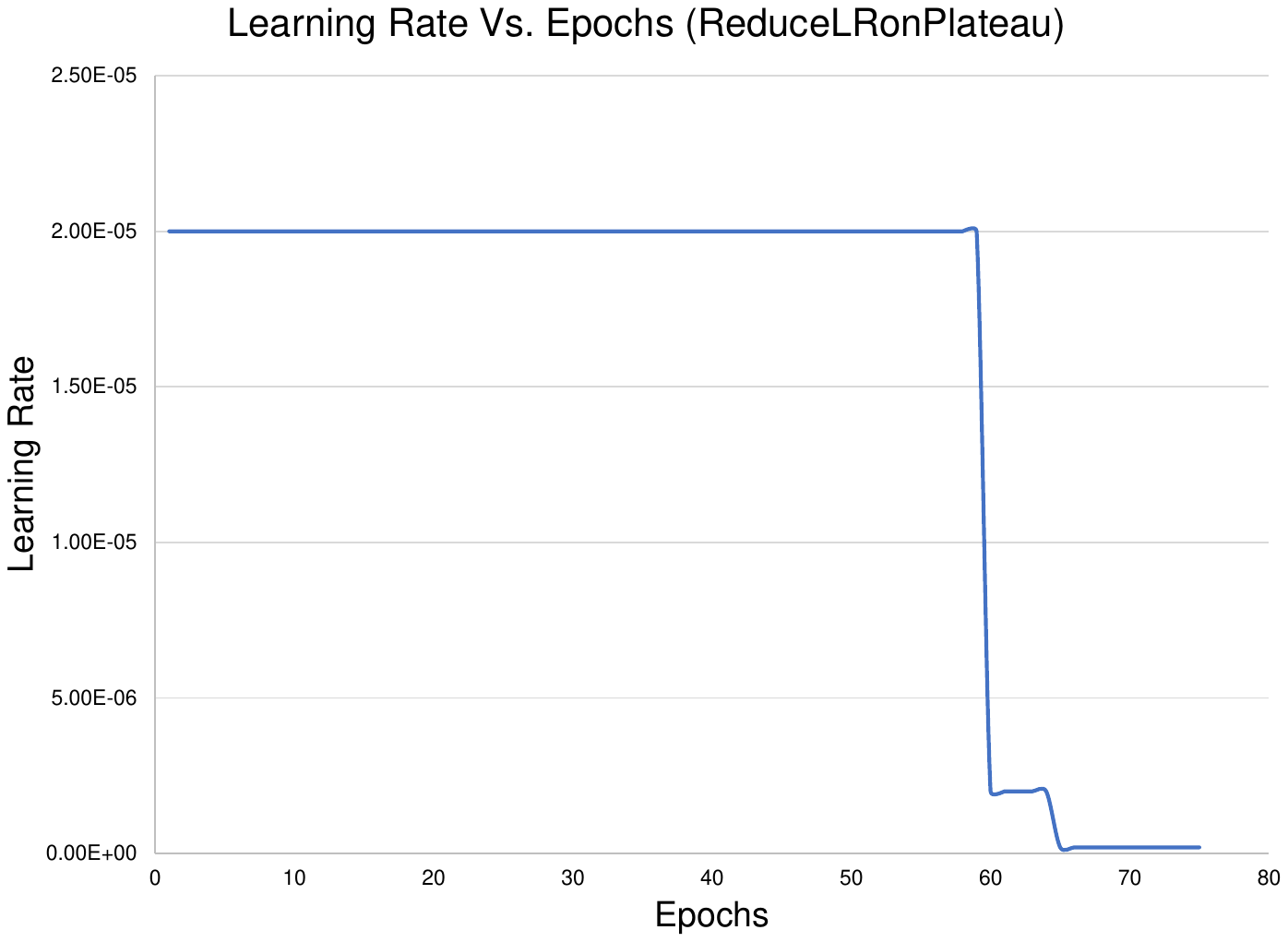}
    \caption{}
    \label{fig:schedulers-behavior-b}
    \end{subfigure}
    \caption{LR change via (a) OneCycleLR and (b) ReduceLRonPlateau.}
    \label{fig:schedulers-behavior}
\end{figure}

Figure~\ref{fig:schedulers-behavior} shows the LR values manipulation during the training phase (assuming that the early stoppage does not intervene). Both are available in the PyTorch library, allowing easy deployment to test their efficacy with the liver segmentation task.

Table~\ref{tab:schedulers-comparison} reports the mean (standard deviation) from the 5-fold cross-validation implemented per scheduler and LR, which allows for understanding the overall generic behavior and the intrinsic difference between each fold run. More details of this part can be found in \cite{alkababji2022scheduling}.

Table~\ref{tab:schedulers-comparison} highlights the top-performing metrics in bold. The results, utilizing the original U-Net in both 2D and 2.5D configurations, surpass those detailed in \cite{Tian2019}, particularly in achieving superior Dice scores compared to the 2D (with and without convolutional long short-term memory (LSTM)), the 2.5D, and the 3D U-Net variants for liver segmentation tasks. Notably, these scores approach closely to those of the GLC-UNet, which reports a Dice score of \(98.18 \pm 0.85\%\). The highest performing instance using ReduceLRonPlateau, with an LR setting of \(16 \times 10^{-5}\), achieves a Dice score of \(98.12 \pm 0.04\%\). This suggests potential improvements in GLC-UNet performance with the adoption of the suggested preprocessing and scheduling strategies. Additionally, the consistency across different test folds is indicated by the minimal variation in results, as evidenced by the small reported standard deviations.

Furthermore, the analysis reveals that selecting an appropriately high LR is crucial to avoid prolonged training durations and suboptimal validation performance due to the network getting stuck at a local minimum. This principle is validated by observations that higher LRs facilitate better generalization of ConvNets in shorter training spans. A comparison of the two scheduling techniques shows that the best performance from ReduceLRonPlateau (LR = \(16 \times 10^{-5}\) with Dice = \(98.12 \pm 0.04\%\)) outperforms the optimal result from OneCycleLR (LRmax = \(24 \times 10^{-5}\) with Dice = \(98.07 \pm 0.11\%\)). Additionally, ReduceLRonPlateau exhibits quicker convergence compared to OneCycleLR, which starts with a lower initial LR that incrementally increases before decreasing.

\begin{table}[!ht]
    \centering
    \begin{tabular}{
>{\centering\arraybackslash}m{0.08\textwidth}
>{\centering\arraybackslash}m{0.07\textwidth}
>{\centering\arraybackslash}m{0.07\textwidth}
>{\centering\arraybackslash}m{0.07\textwidth}
>{\centering\arraybackslash}m{0.07\textwidth}
>{\centering\arraybackslash}m{0.07\textwidth}
>{\centering\arraybackslash}m{0.07\textwidth}
>{\centering\arraybackslash}m{0.07\textwidth}
>{\centering\arraybackslash}m{0.07\textwidth}
>{\centering\arraybackslash}m{0.07\textwidth}
}
\hline
\textbf{Scheduler}                              & \textbf{LR ($\times10^{-5}$)} & \textbf{Dice (\%)}    & \textbf{IoU (\%)}     & \textbf{RVD}            & \textbf{ASD (mm)}      & \textbf{RMSD (mm)}   & \textbf{HD (mm)}       & \textbf{95\% HD (mm)} & \textbf{Epochs} \\ \hline
\multirow{13}{*}{\rotatebox[origin=c]{90}{OneCycleLR (Max LR)}} 
    & 2         & 97.67 (0.12) & 95.48 (0.23) & -0.008 (0.004) & 1.055 (0.509) & 3.81 (1.15) & 44.00 (4.43)  & 5.13 (4.03)  & 42.6 (2.6) \\
    & 4         & 97.72 (0.08) & 95.58 (0.28) & -0.009 (0.004) & 1.002 (0.522) & 3.70 (1.58) & 41.58 (7.41)  & 5.09 (4.12)  & 34.0 (3.1) \\
    & 8         & 97.93 (0.09) & 95.97 (0.17) & -0.009 (0.005) & 0.576 (0.089) & 2.51 (0.30) & 37.78 (1.86)  & 2.82 (0.70)  & 29.4 (2.2) \\
    & 16        & 98.01 (0.06) & 96.12 (0.11) & -0.008 (0.005) & 0.521 (0.110) & 2.05 (0.45) & 31.51 (6.71)  & 2.68 (0.67)  & 29.0 (2.9) \\
    & 24        & 98.07 (0.11) & 96.24 (0.21) & -0.005 (0.003) & \textbf{0.482 (0.076)} & \textbf{1.88 (0.29)} & 29.25 (5.18)  & \textbf{2.43 (0.59)}  & 30.0 (3.8) \\
    & 32        & 98.03 (0.10) & 96.16 (0.19) & \textbf{-0.004 (0.005)} & 0.606 (0.147) & 2.47 (0.71) & 34.31 (3.11)  & 2.61 (0.45)  & 25.0 (3.7) \\
    & 40        & 98.04 (0.17) & 96.17 (0.32) & -0.007 (0.003) & 0.503 (0.131) & 1.97 (0.74) & 33.17 (8.63)  & 2.56 (0.82)  & 25.8 (2.0) \\ \hline
\multirow{13}{*}{\rotatebox[origin=c]{90}{ReduceLRonPlateau}}     
    & 0.4       & 97.20 (0.13) & 94.60 (0.24) & -0.009 (0.004) & 1.541 (0.276) & 5.59 (0.63) & 59.02 (7.91)  & 10.63 (3.15) & 72.8 (3.9) \\
    & 0.8       & 97.40 (0.08) & 94.96 (0.27) & -0.010 (0.003) & 1.097 (0.238) & 4.47 (0.86) & 59.47 (7.83)  & 5.56 (3.35)  & 48.6 (4.0) \\
    & 2         & 97.69 (0.13) & 95.51 (0.24) & -0.011 (0.003) & 0.751 (0.130) & 3.03 (0.61) & 40.43 (1.46)  & 3.40 (0.76)  & 30.6 (1.3) \\
    & 4         & 97.87 (0.06) & 95.86 (0.11) & -0.011 (0.003) & 0.625 (0.142) & 2.57 (0.84) & 38.26 (7.34)  & 2.73 (0.12)  & 23.0 (2.8) \\
    & 8         & 98.04 (0.10) & 96.18 (0.19) & -0.009 (0.003) & 0.528 (0.104) & 1.99 (0.52) & \textbf{27.03 (6.55)}  & 2.67 (0.73)  & 20.8 (3.6) \\
    & 16        & \textbf{98.12 (0.04)} & \textbf{96.33 (0.07)} & -0.008 (0.002) & 0.624 (0.443) & 2.15 (1.40) & 27.16 (4.53)  & 4.10 (4.37)  & 22.0 (1.6) \\
    & 32        & 98.02 (0.10) & 96.15 (0.19) & -0.006 (0.004) & 0.515 (0.062) & 2.09 (0.52) & 33.87 (10.39) & 2.58 (0.42)  & \textbf{18.8 (4.2)} \\ \hline
\end{tabular}

    \caption{Both OneCycleLR and ReduceLRonPlateau schedulers with different LRs are reported for each metric.}
    \label{tab:schedulers-comparison}
\end{table}

\subsection{Tumors and Vessels Segmentation Challenge}

Ground-truth masks for tumors and vessels are available for 14 of the 23 records in the testing dataset, specifically identified as \textbf{[072, 090, 117, 129, 141, 178, 193, 236, 246, 258, 268, 280, 294, 320]}. This dataset subset provides a robust base for evaluating the segmentation ConvNets designed for tumors and vessels. During the implementation of 5-fold cross-validation, approximately 12,000 slices are allocated to the four training folds, and about 3,000 slices are designated for the validation fold in the fifth segment.

\begin{figure}[!ht]
    \centering
    \includegraphics{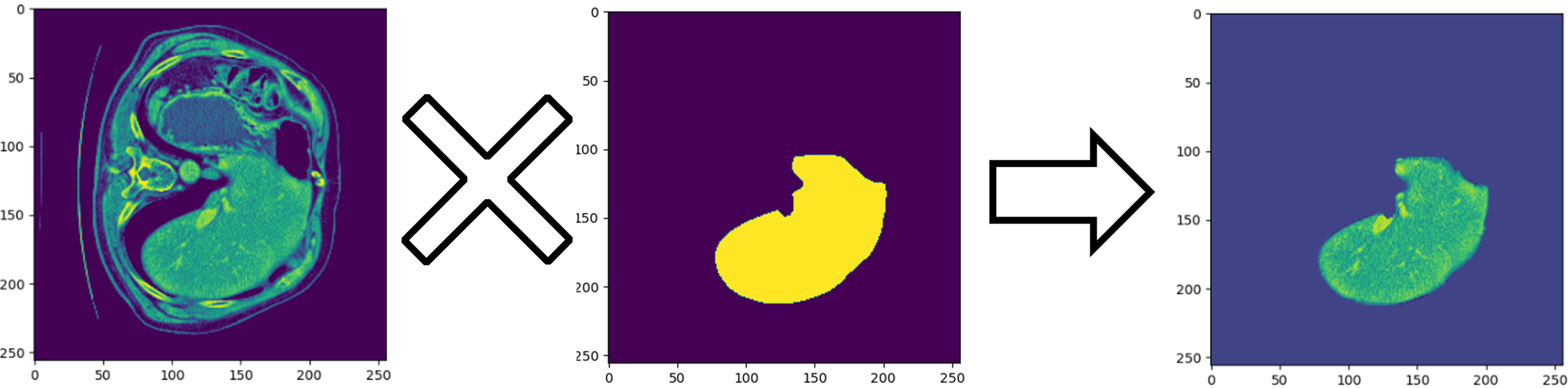}
    \caption{Element-wise multiplication of the CT scan slice with its liver mask.}
    \label{fig:background-voxels-removal}
\end{figure}

Upon identifying the optimal liver segmentation model, the approach extends to address segmentation of tumors and vessels. Notably, segmenting the liver first is beneficial for refining the focus of the subsequent tumor and vessel ConvNets. This preprocessing step involves the elimination of background voxels, enhancing the focus on the liver's critical regions, as illustrated in Figure~\ref{fig:overall-system}. An example of this process is demonstrated in Figure~\ref{fig:background-voxels-removal}, where an element-wise multiplication is performed between a CT scan slice (record \#369, slice 128) and its corresponding ground-truth mask to produce segmented slices that serve as input for subsequent ConvNets.

Following the initial liver segmentation by the first ConvNet, the subsequent tumor and vessel ConvNets utilize the segmented liver output as a mask. This approach helps focus analysis on liver voxels and effectively ignores non-liver voxels, treating them as background. To assess the impact of this method, Table~\ref{tab:before-after-liver-mask-multiplication} presents comparative results before and after the application of the liver mask in tumor and vessel segmentation tasks. In this evaluation, the OneCycleLR scheduler is employed, as it has shown superior performance over the ReduceLRonPlateau scheduler in these specific segmentation challenges.

\begin{table}[!ht]
    \centering
    \caption{Performance metrics of both tumor and vessel ConvNets before and after applying liver masks.}
    \begin{tabular}{
>{\centering\arraybackslash}m{0.05\textwidth}
>{\centering\arraybackslash}m{0.08\textwidth}
>{\centering\arraybackslash}m{0.08\textwidth}
>{\centering\arraybackslash}m{0.08\textwidth}
>{\centering\arraybackslash}m{0.08\textwidth}
>{\centering\arraybackslash}m{0.08\textwidth}
>{\centering\arraybackslash}m{0.08\textwidth}
>{\centering\arraybackslash}m{0.08\textwidth}
>{\centering\arraybackslash}m{0.08\textwidth}
>{\centering\arraybackslash}m{0.08\textwidth}
} \hline
TOI                      & Stage                   & LR ($\times10^{-5}$) & Dice (\%)             & IoU (\%)              & RVD                    & ASD (mm)                & RMSD (mm)             & HD (mm)               & 95\% HD (mm)           \\ \hline
\multirow{11}{*}{\rotatebox[origin=c]{90}{Tumors}}  & \multirow{5}{*}{Before} & 4         & 38.62 (3.54)          & 27.66 (2.78)          & -0.335 (0.108)         & 38.018 (12.247)         & 46.08 (12.5)          & 105.37 (13.59)        & 71.46 (16.35)          \\
    &                         & 28        & 51.18 (4.08)          & 39.09 (3.72)          & -0.042 (0.247)         & 41.948 (13.383)         & 51.53 (12.87)         & 120.27 (20.09)        & 76.85 (18.39)          \\
    &                         & 40        & 53.24 (4.03)          & 40.53 (3.88)          & 0.252 (0.371)          & 31.109 (11.35)          & 41.67 (9.68)          & 120.55 (17.12)        & 69.11 (10.34)          \\ \cline{2-10}
    & \multirow{5}{*}{After}  & 4         & 57.73 (1.1)           & 45.19 (0.86)          & \textbf{0.040 (0.162)} & 25.299 (16.011)         & 32.15 (15.2)          & 89.57 (11.13)         & 51.03 (14.22)          \\
    &                         & 28        & \textbf{59.71 (3.91)} & \textbf{46.94 (3.68)} & 0.251 (0.328)          & 17.126 (13.208)         & 23.25 (12.18)         & 77.39 (8.22)          & \textbf{38.49 (11.01)} \\
    &                         & 40        & 57.86 (2.41)          & 45.22 (2.65)          & 0.312 (0.514)          & \textbf{11.445 (1.618)} & \textbf{18.04 (1.73)} & \textbf{68.95 (5.63)} & 38.57 (5.92)           \\ \hline
\multirow{11}{*}{\rotatebox[origin=c]{90}{Vessels}} & \multirow{5}{*}{Before} & 4         & 50.50 (1.18)          & 34.79 (1.07)          & 0.052 (0.118)          & 4.363 (0.336)           & 9.15 (0.57)           & 79.64 (10.09)         & 19.06 (1.46)           \\
 &                         & 28        & \textbf{52.68 (1.38)} & \textbf{36.85 (1.06)} & 0.065 (0.094)          & 4.563 (0.376)           & 10.27 (1.98)          & 72.09 (17.79)         & 19.81 (0.87)           \\
 &                         & 40        & 52.39 (1.84)          & 36.63 (1.74)          & \textbf{0.038 (0.115)} & \textbf{4.164 (0.163)}  & 8.22 (0.21)           & 58.13 (13.10)         & 18.51 (0.87)           \\ \cline{2-10}
 & \multirow{5}{*}{After}  & 4         & 48.63 (1.30)          & 33.18 (1.13)          & -0.098 (0.133)         & 4.230 (0.395)           & 7.97 (0.54)           & 49.25 (6.05)          & \textbf{18.34 (1.33)}  \\
 &                         & 28        & 50.34 (1.82)          & 34.73 (1.52)          & -0.045 (0.144)         & 4.186 (0.350)           & \textbf{7.95 (0.44)}  & 44.70 (3.64)          & 18.96 (1.30)           \\
 &                         & 40        & 47.36 (1.85)          & 32.28 (1.57)          & -0.244 (0.065)         & 4.800 (0.473)           & 8.61 (0.56)           & \textbf{42.88 (1.34)} & 20.46 (1.23) \\ \hline
\end{tabular}
    \label{tab:before-after-liver-mask-multiplication}
\end{table}

\begin{figure}[!ht]
    \centering
    \includegraphics[width=1\linewidth]{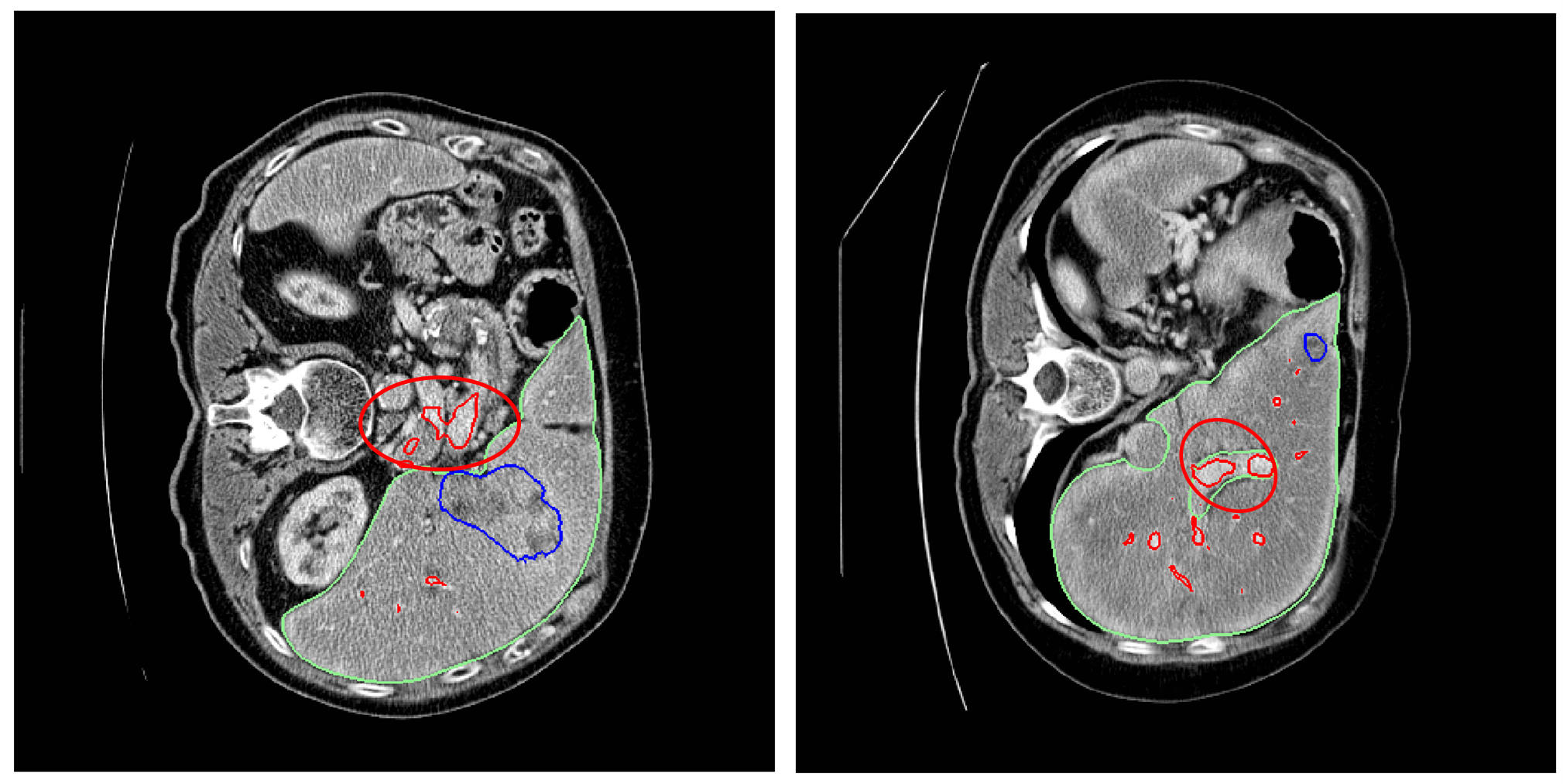}
    \begin{subfigure}[b]{0.49\textwidth}
    \caption{Hepaticvessel\_294\_slice\_83}
    \label{fig:vessels-location-a}
    \end{subfigure}
    \hfill
    \begin{subfigure}[b]{0.49\textwidth}
    \caption{Hepaticvessel\_268\_slice\_27}
    \label{fig:vessels-location-b}
    \end{subfigure}
    \caption{Illustrations of vasculature tissues are presented, where illustration (a) depicts vessels entirely external to the liver, and illustration (b) displays vessels that are within the liver organ yet outside the designated liver mask.
.}
    \label{fig:vessels-location}
\end{figure}

The application of the liver mask markedly enhanced the segmentation performance for tumors but did not yield similar improvements for vessels; in fact, vessel segmentation performance marginally declined. This disparity can be explained by two main factors: Firstly, the liver's vessel structure possesses distinctive characteristics that make the liver mask less beneficial for vessel segmentation compared to tumor segmentation. Tumors, which can be mistaken for other organ tissues such as the kidneys, benefit from the application of the liver mask as it helps to exclude these non-tumor tissues. Secondly, the construction of liver masks sometimes excludes vessels. Consequently, when the liver mask is applied, it may inadvertently eliminate vessel voxels. Figure~\ref{fig:vessels-location} illustrates this issue with two examples where vascular tissues are omitted following the application of the liver mask. In these examples, green represents the ground-truth liver mask, blue the ground-truth tumors mask, and red the ground-truth vessels mask.

Considering the advantages and challenges of applying liver delineation to subsequent ConvNets, liver masks are consistently used for both tumor and vessel tissues, with plans to modify the approach for vessel tissues in future studies. The learning rates (LRs) employed for tumor and vessel segmentation tasks are detailed in Tables~\ref{tab:convnets-tumors} and \ref{tab:convnets-vessels}, respectively. These tables provide the mean and standard deviation from five runs of 5-fold cross-validation.

For tumor segmentation, as illustrated in Table~\ref{tab:convnets-tumors}, variations in LRs did not significantly enhance the performance metrics, although their effects are more evident in Table~\ref{tab:before-after-liver-mask-multiplication}. Notably, while the Dice scores for LRs of \(28 \times 10^{-5}\) and \(150 \times 10^{-5}\) are comparable, the standard deviation is considerably reduced at \(150 \times 10^{-5}\), indicating more consistent model behavior at this higher LR.

\begin{table}[!ht]
    \centering
    \caption{Performance evaluation of tumor ConvNets with different LRs.}
    \begin{tabular}{
>{\centering\arraybackslash}m{0.08\textwidth}
>{\centering\arraybackslash}m{0.08\textwidth}
>{\centering\arraybackslash}m{0.08\textwidth}
>{\centering\arraybackslash}m{0.08\textwidth}
>{\centering\arraybackslash}m{0.08\textwidth}
>{\centering\arraybackslash}m{0.08\textwidth}
>{\centering\arraybackslash}m{0.08\textwidth}
>{\centering\arraybackslash}m{0.08\textwidth}
}
\toprule
LR ($\times10^{-5}$) & Dice (\%)             & IoU (\%)              & RVD                    & ASD (mm)                & RMSD (mm)             & HD (mm)               & 95\% HD (mm)          \\ \midrule
4         & 57.73 (1.10)          & 45.19 (0.86)          & \textbf{0.040 (0.162)} & 25.299 (16.011)         & 32.15 (15.20)         & 89.57 (11.13)         & 51.03 (14.22)         \\
20        & 59.02 (3.64)          & 46.79 (3.73)          & 0.174 (0.231)          & 24.834 (15.989)         & 31.23 (15.40)         & 80.61 (14.21)         & 50.58 (15.24)         \\
28        & \textbf{59.71 (3.91)} & 46.94 (3.68)          & 0.251 (0.328)          & 17.126 (13.208)         & 23.25 (12.18)         & 77.39 (8.22)          & 38.49 (11.01)         \\
40        & 57.86 (2.41)          & 45.22 (2.65)          & 0.312 (0.514)          & \textbf{11.445 (1.618)} & 18.04 (1.73)          & \textbf{68.95 (5.63)} & 38.57 (5.92)          \\
60        & 57.63 (1.63)          & 45.18 (2.12)          & 0.083 (0.228)          & 23.249 (23.505)         & 28.98 (23.79)         & 75.07 (22.5)          & 44.81 (27.03)         \\
80        & 58.75 (2.27)          & 46.38 (1.98)          & 0.310 (0.184)          & 17.490 (13.207)         & 23.94 (12.10)         & 74.02 (11.71)         & 42.82 (11.36)         \\
100       & 57.73 (4.05)          & 45.60 (3.72)          & 0.513 (0.436)          & 11.991 (1.564)          & 18.92 (2.7)           & 73.10 (10.13)         & 39.42 (9.32)          \\
120       & 59.51 (3.00)          & 47.08 (3.49)          & 0.467 (0.431)          & 11.954 (2.613)          & \textbf{17.92 (2.99)} & 72.46 (7.52)          & \textbf{34.43 (3.84)} \\
150       & 59.70 (0.61)          & \textbf{47.28 (1.24)} & 0.229 (0.263)          & 11.921 (0.988)          & 19.04 (2.15)          & 73.18 (8.83)          & 38.82 (6.59)          \\ \bottomrule
\end{tabular}
    \label{tab:convnets-tumors}
\end{table}

\begin{table}[!ht]
    \centering
    \caption{Performance evaluation of vessel ConvNets with different LRs.}
    \begin{tabular}{
>{\centering\arraybackslash}m{0.08\textwidth}
>{\centering\arraybackslash}m{0.08\textwidth}
>{\centering\arraybackslash}m{0.08\textwidth}
>{\centering\arraybackslash}m{0.08\textwidth}
>{\centering\arraybackslash}m{0.08\textwidth}
>{\centering\arraybackslash}m{0.08\textwidth}
>{\centering\arraybackslash}m{0.08\textwidth}
>{\centering\arraybackslash}m{0.08\textwidth}
}
\toprule
LR ($\times10^{-5}$) & Dice (\%)             & IoU (\%)              & RVD                     & ASD (mm)               & RMSD (mm)            & HD (mm)               & 95\% HD (mm)          \\ \midrule
4         & 48.63 (1.30)          & 33.18 (1.13)          & -0.098 (0.133)          & 4.230 (0.395)          & 7.97 (0.54)          & 49.25 (6.05)          & \textbf{18.34 (1.33)} \\
20        & 48.39 (2.19)          & 33.04 (1.81)          & -0.095 (0.092)          & 5.091 (1.157)          & 9.15 (1.47)          & 46.59 (2.93)          & 21.40 (2.83)          \\
28        & \textbf{50.34 (1.82)} & \textbf{34.73 (1.52)} & \textbf{-0.045 (0.144)} & \textbf{4.186 (0.350)} & \textbf{7.95 (0.44)} & 44.70 (3.64)          & 18.96 (1.30)          \\
40        & 47.36 (1.85)          & 32.28 (1.57)          & -0.244 (0.065)          & 4.800 (0.473)          & 8.61 (0.56)          & \textbf{42.88 (1.34)} & 20.46 (1.23)          \\
60        & 49.72 (2.40)          & 34.34 (2.01)          & -0.196 (0.119)          & 4.422 (0.732)          & 8.13 (0.95)          & 43.67 (2.12)          & 19.29 (2.01)          \\
80        & 48.95 (3.01)          & 33.61 (2.63)          & -0.150 (0.144)          & 4.559 (0.764)          & 8.35 (0.86)          & 46.80 (4.10)          & 19.59 (2.20)          \\
100       & 47.66 (0.96)          & 32.39 (0.89)          & -0.277 (0.102)          & 4.900 (0.304)          & 8.95 (0.50)          & 44.64 (3.01)          & 21.33 (1.09)          \\
120       & 49.03 (4.02)          & 33.62 (3.36)          & -0.183 (0.259)          & 4.483 (0.757)          & 8.23 (0.82)          & 43.68 (5.04)          & 19.37 (1.59)          \\
150       & 48.58 (1.87)          & 33.30 (1.61)          & -0.239 (0.142)          & 4.843 (0.786)          & 8.68 (0.95)          & 44.36 (1.64)          & 20.57 (1.84)          \\ \bottomrule
\end{tabular}
    \label{tab:convnets-vessels}
\end{table}

Overall, in vessel segmentation, modifications in LR showed minimal impact on both the performance metrics and their standard deviations, suggesting a consistent performance across different LRs. These findings are in line with those reported on the MSDC challenge website for the Task 8 Hepatic Vessel challenge. The challenge results reflect a multi-class segmentation scenario, which, while potentially simplifying the differentiation between tumor and vessel tissues, also adds the complexity of multi-organ segmentation.

\subsection{Best ConvNets Results Showcase}
In this subsection, we briefly mention the used framework, programming language, training parameters and environment, and the best ConvNet models achieved during our 5-fold cross-validation trials and the other folds existing in those trials. It is worth noting that multiple metrics have been used to evaluate the performance of the trained models. We used the Python programming language to conduct our work for both front-end and back-end. It has a plethora of existing modules designed to make the development of ML projects and GUI applications. For the ML part, we have opted for the PyTorch framework to build the ConvNets. PyTorch (version=torch version==1.8.1+cu102) is used, along with scikit-learn package to create the 5-fold cross-validation procedure (80\% training/20\% validation). The defined maximum number of epochs is 75 per fold, but early stopping intervenes when the validation loss does not improve with epochs\_stop~=~6. Moreover, 2.5D (5 slices) as an input shape is utilized with a batch size~=~32, and Adam optimizer is used to optimize the ConvNet with respect to the binary cross-entropy (BCE) loss function. ReduceLRonPlateau~\cite{PyTorch-reducelronplateau} is used for training the liver segmentation ConvNet, while OneCycleLR~\cite{PyTorch-onecyclelr} scheduler is employed for the tumors and vessels segmentation ConvNets. The torch.optim module in PyTorch facilitates creating and deploying both schedulers. Table~\ref{tab:convnets} shows the best-achieved trials for the three tissues (liver, tumors, and vessels). It can be noticed that the best fold for liver segmentation is either fold \#2 or \#5 (we have opted for \#2), fold \#2 for the tumor segmentation, and fold \#1 for the vessel segmentation challenge.

\begin{table}[!ht]
\centering
\caption{Best U-Net ConvNets' trials 5-fold cross-validation with learning rate = 16$\times$10\textsuperscript{-5} for liver, and learning rate = 28$\times$10\textsuperscript{-5} for both tumors and vessels ConvNets}
\resizebox{\linewidth}{!}{%
\begin{tabular}{
>{\centering\arraybackslash}m{1cm}
>{\centering\arraybackslash}m{0.9cm}
>{\centering\arraybackslash}m{1.5cm}
>{\centering\arraybackslash}m{1.5cm}
>{\centering\arraybackslash}m{1.5cm}
>{\centering\arraybackslash}m{1.5cm}
>{\centering\arraybackslash}m{1.5cm}
>{\centering\arraybackslash}m{1.5cm}
>{\centering\arraybackslash}m{2.5cm}
}
\hline
\textbf{Tissue}          & \textbf{Fold}                 & \textbf{Dice (\%)} & \textbf{IoU (\%)} & \textbf{RVD}    & \textbf{ASD (mm)} & \textbf{RMSD (mm)} & \textbf{HD (mm)} & \textbf{95\% HD (HD95) (mm)} \\ \hline
\multirow{7}{*}{Liver} 
    & 1                             & 98.09              & 96.27             & -0.008          & 1.41              & 4.59               & 31.70            & 11.92                        \\
    & \textbf{2}                    & \textbf{98.16}     & \textbf{96.41}    & -0.011          & 0.41              & 1.44               & 25.69            & 2.14                         \\
    & 3                             & 98.12              & 96.33             & \textbf{-0.005} & \textbf{0.39}     & \textbf{1.29}      & 31.36            & \textbf{2.11}                \\
    & 4                             & 98.08              & 96.26             & -0.008          & 0.51              & 2.07               & 26.29            & 2.19                         \\
    & 5                             & \textbf{98.16}     & \textbf{96.41}    & -0.009          & 0.40              & 1.37               & \textbf{20.75}   & 2.16                         \\
    & Mean (SD)                     & 98.12 (0.04)       & 96.33 (0.07)      & -0.008 (0.002)  & 0.62 (0.44)       & 2.15 \quad\quad\quad(1.4)         & 27.16 (4.53)     & 4.1 \quad\quad\quad\quad\quad(4.37)             \\ \hline
\multirow{7}{*}{Tumors} 
    & 1                             & 56.77              & 43.98             & 0.664           & 12.12             & 18.50              & 79.30            & 40.94                        \\
    & \textbf{2}                    & \textbf{65.95}     & \textbf{52.92}    & \textbf{0.091}  & \textbf{9.06}     & \textbf{14.84}     & \textbf{63.68}   & \textbf{26.00}               \\
    & 3                             & 56.28              & 43.94             & -0.207          & 40.64             & 44.79              & 82.71            & 55.66                        \\
    & 4                             & 60.75              & 47.50             & 0.399           & 12.37             & 19.82              & 84.48            & 36.15                        \\
    & 5                             & 58.79              & 46.36             & 0.309           & 11.44             & 18.29              & 76.79            & 33.70                        \\
    & Mean (SD) & 59.71 (3.91)       & 46.94 (3.68)      & 0.251 (0.328)   & 17.13 (13.21)     & 23.25 (12.18)      & 77.39 (8.22)     & 38.49 \quad\quad\quad\quad(11.01)            \\ \hline
\multirow{7}{*}{Vessels} & 
\textbf{1}                    & \textbf{51.94}     & \textbf{36.11}    & 0.078           & \textbf{3.85}     & \textbf{7.53}      & 45.33            & 17.90                        \\
    & 2                             & 50.81              & 35.19             & 0.084           & 4.05              & 7.85               & 50.02            & 18.34                        \\
    & 3                             & 51.32              & 35.58             & -0.041          & 4.25              & 8.31               & 45.48            & 20.70                        \\
    & 4                             & 47.26              & 32.21             & -0.269          & 4.76              & 8.51               & 41.98            & 20.01                        \\
    & 5                             & 50.34              & 34.57             & \textbf{-0.076} & 4.03              & 7.58               & \textbf{40.69}   & \textbf{17.88}               \\
    & Mean (SD) & 50.34 (1.82)       & 34.73 (1.52)      & -0.045 (0.144)  & 4.186 (0.35)      & 7.95 (0.44)        & 44.7 (3.64)      & 18.96 \quad\quad\quad\quad(1.3)              \\ \hline
\end{tabular}}
\label{tab:convnets}
\end{table}

Note that the used scheduler for the liver ConvNet is the ReduceLRonPlateau, while the tumors and vessels ConvNets utilized the OneCycleLR scheduler.

\begin{figure}[!ht]
    \centering
    \includegraphics[width=0.8\linewidth]{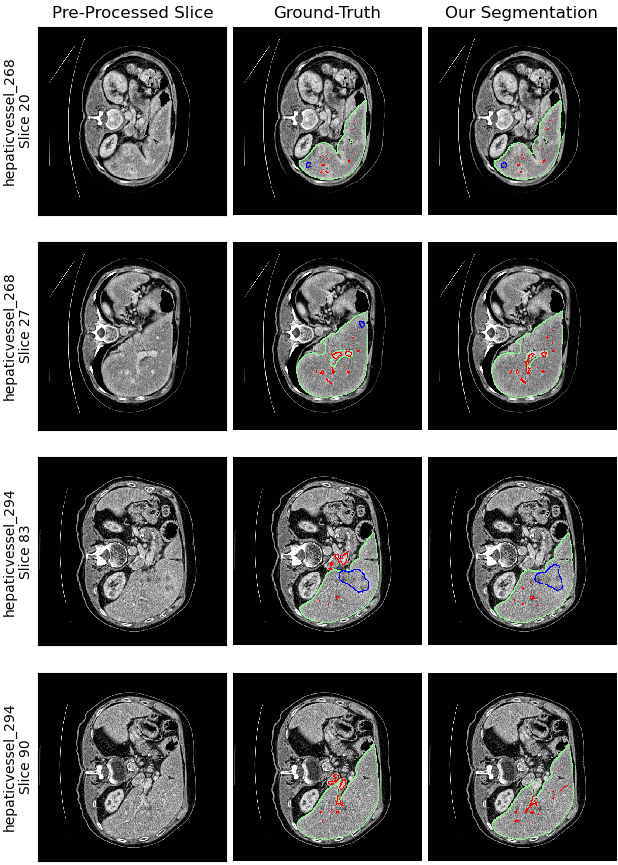}
    \caption{A comparison between the ground-truth masks our models' performance. Green, blue, and red highlight the liver, tumors, and vessel masks' borders, respectively.}
    \label{fig:2d-segmentation-showcase}
\end{figure}

Figure~\ref{fig:2d-segmentation-showcase} demonstrates a comparison between \enquote{Our Segmentation}, representing the combined results of the three optimally trained ConvNets, and the \enquote{Ground-Truth} masks. Specifically, for volume hepaticvessel\_268, the liver segmentation exhibits near-perfect alignment in slices 20 and 27.

For the liver parenchyma segmentation, we can confidently mention that the liver ConvNet has successfully segmented the liver, reaching ground-truth precision.



The ConvNet's performance in tumor segmentation varies across different slices: it performs well in slice 20, yet fails to detect a small tumor in slice 27. Conversely, in the case of vessel detection, while the ConvNet overlooks some minor vessels in slice 20, it excels in slices 83 and 90, capturing all the vessels effectively---surpassing even the ground-truth mask in accuracy.

Focusing on slice 83 within the hepaticvessel\_294 volume, the tumor segmentation ConvNet successfully identifies and accurately delineates the tumor, demonstrating high recall. Similarly, the vessel ConvNet effectively captures all vessel tissues within the liver, suggesting its performance might exceed that of the original mask. However, vessels located outside the liver are not detected, a limitation imposed by the application of the liver mask. This observation highlights a notable insight: the impact of the liver mask on vessel detection varies based on their proximity to the liver. For instance, while vessels within the liver but outside the liver mask in hepaticvessel\_268 are detected, those in hepaticvessel\_294 that lie outside both the liver and its mask are missed. This differential effect underscores the complex interaction between the segmentation process and the anatomical positioning of the vessels.
This could be attributed to the inter-variance of the dataset that we have, where some liver masks would have considered the inner vessels to be part of the liver, while others, such as the case of hepaticvessel\_268, consider vessels outside of the liver mask.

\subsection{AI Radiologist Implementation and Showcase}
\begin{figure}[!ht]
    \centering
    \includegraphics[width=0.75\textwidth]{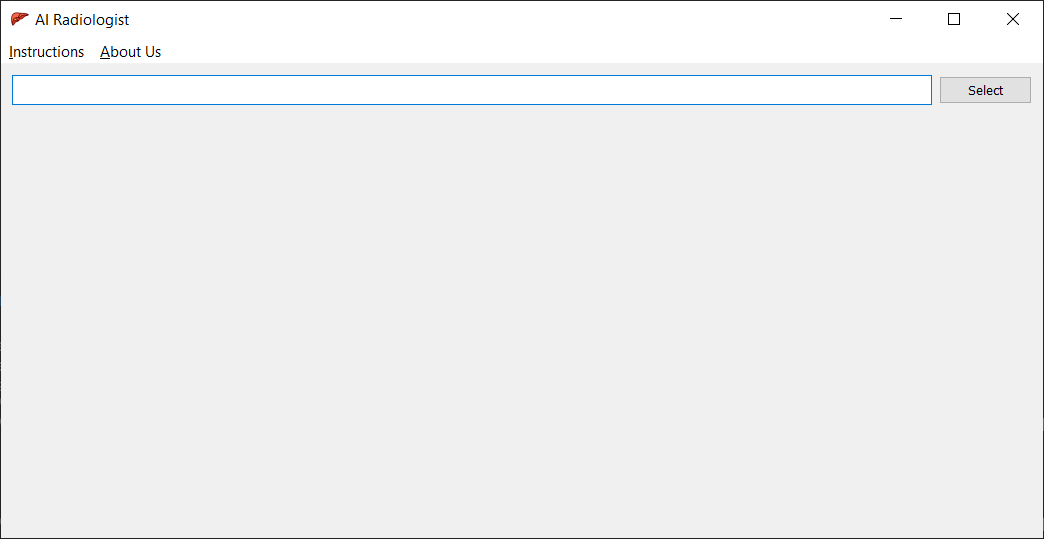}
    \caption{GUI desktop application initial screen.}
    \label{fig:desktop_application}
\end{figure}

\begin{figure}[!ht]
    \centering
    \includegraphics[width=0.75\textwidth]{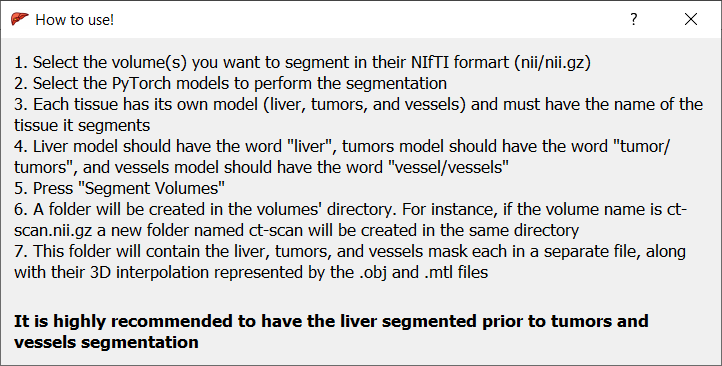}
    \caption{AI Radiologist usage instructions.}
    \label{fig:usage_instructions}
\end{figure}

This section presents the subsequent steps following the development of three ConvNet models designed to segment liver parenchyma, tumors, and vessels. We utilize the PyQt5 framework~\cite{PyQt5} to construct a desktop application in Python. Figure~\ref{fig:desktop_application} displays the main interface that greets users upon launching the application, named "AI Radiologist" and symbolized by a liver icon. Below this, the Instructions menu tab provides a guide on using the application, as illustrated in Figure~\ref{fig:usage_instructions}.

\begin{figure}[!ht]
    \centering
    \includegraphics[width=0.75\textwidth]{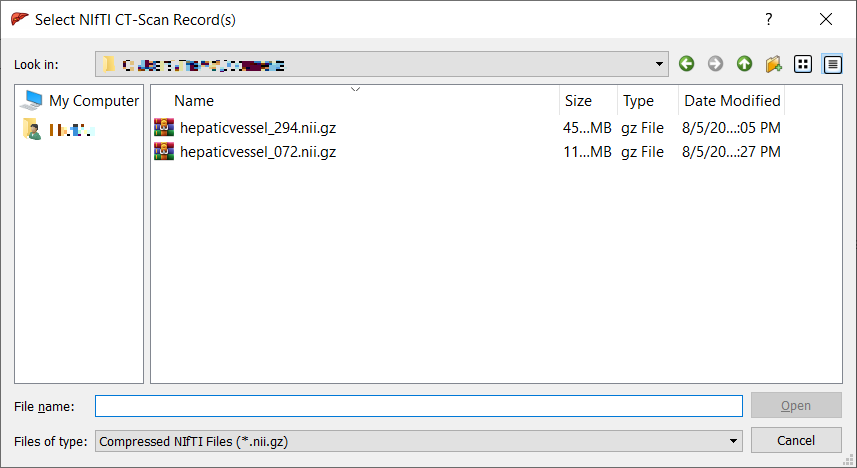}
    \caption{A pop-up window for NIfTI CT scans selection.}
    \label{fig:selecting_records}
\end{figure}

Pressing the \enquote{Select} button allows clinicians to select multiple CT scans (saved in NIfTI format) that are available locally on their devices. Figure~\ref{fig:selecting_records} shows an example of how the files that can be imported, preventing any other files from being imported since they need to be either nii/nii.gz.

\begin{figure}[!hbtp]
    \centering
    \includegraphics[width=0.75\textwidth]{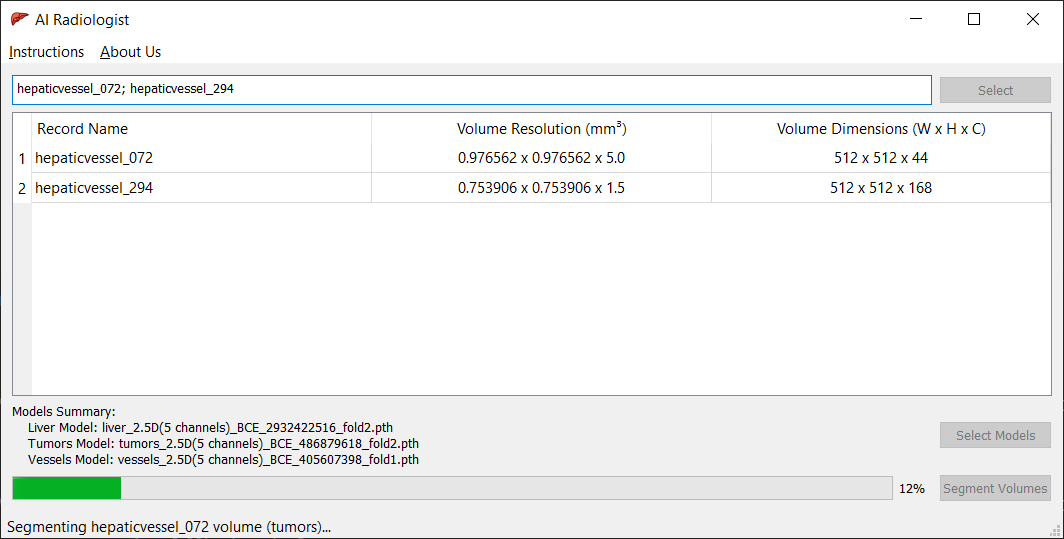}
    \caption{Volumes and models selection and current segmentation progress for processd volumes.}
    \label{fig:application_progressing}
\end{figure}

Once the end-users select the CT scans they want to segment the liver and its tissues within, a table displaying the records’ metadata pops up. Following this step, the user must select the trained PyTorch models for segmenting the liver, tumors, and vessels. The application is ready to perform the segmentation once the clinicians completes importing the CT scans and the ConvNet models. Figure~\ref{fig:application_progressing} portrays choosing multiple volumes for delineation and the selected models for segmentation. The volumes are shown in a table for as many CT records imported to the application.


The heavy-backend processes commence once the \enquote{Segment Volumes} button is pressed. Same preprocessing techniques, that are used to train the models, are applied to the imported volumes. Computations are done efficiently and in real-time, since GPU capabilities are being heavily used. The main thread is strictly used for GUI interactions, while heavy computations are delegated to other threads to avoid blocking the main thread. The application allows clinicians to view the current segmentation status for each volume as illustrated by  Figure~\ref{fig:application_progressing}, which show the volume and the phase the application is currently processing. It is of UX best practices to allow users inside information of the application progress so they can plan their time accordingly.

\begin{figure}[!ht]
    \centering
    \includegraphics[width=0.8\textwidth, trim={1in 0in 1.3in 0in},clip]{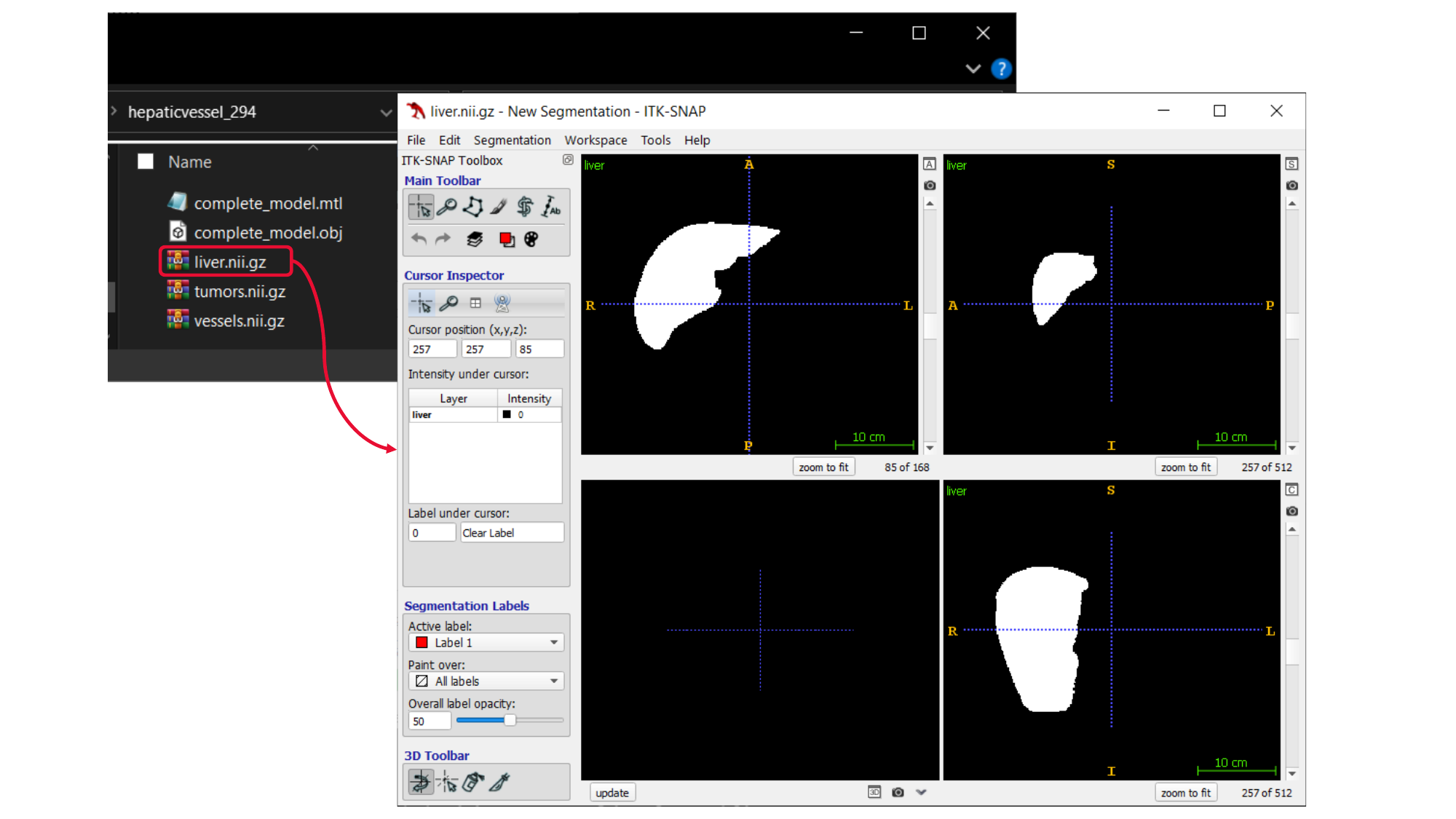}
    \caption{Application's output files for a CT scan, and opening them in a compatible software.}
    \label{fig:app_output}
\end{figure}

After completing the segmentation for the liver tissues and creating the 2D automated masks, the 3D interpolation and merging process commences for the three generated masks. The built 3D object and the constituting masks are placed within the same directory in a folder given the same name as the original CT scan. The application generates the isolated masks in liver.nii.gz, tumors.nii.gz, and vessels.nii.gz for easier access and to allow clinicians the flexibility of viewing these masks in any software they are accustomed to use. Figure~\ref{fig:app_output} illustrates all the resulting files after completely processing volume~294 (MSDC-T8 dataset), and shows how the liver mask is viewed inside the ITK-SNAP software.

\begin{figure}[!ht]
    \centering
    \includegraphics[width=0.8\textwidth]{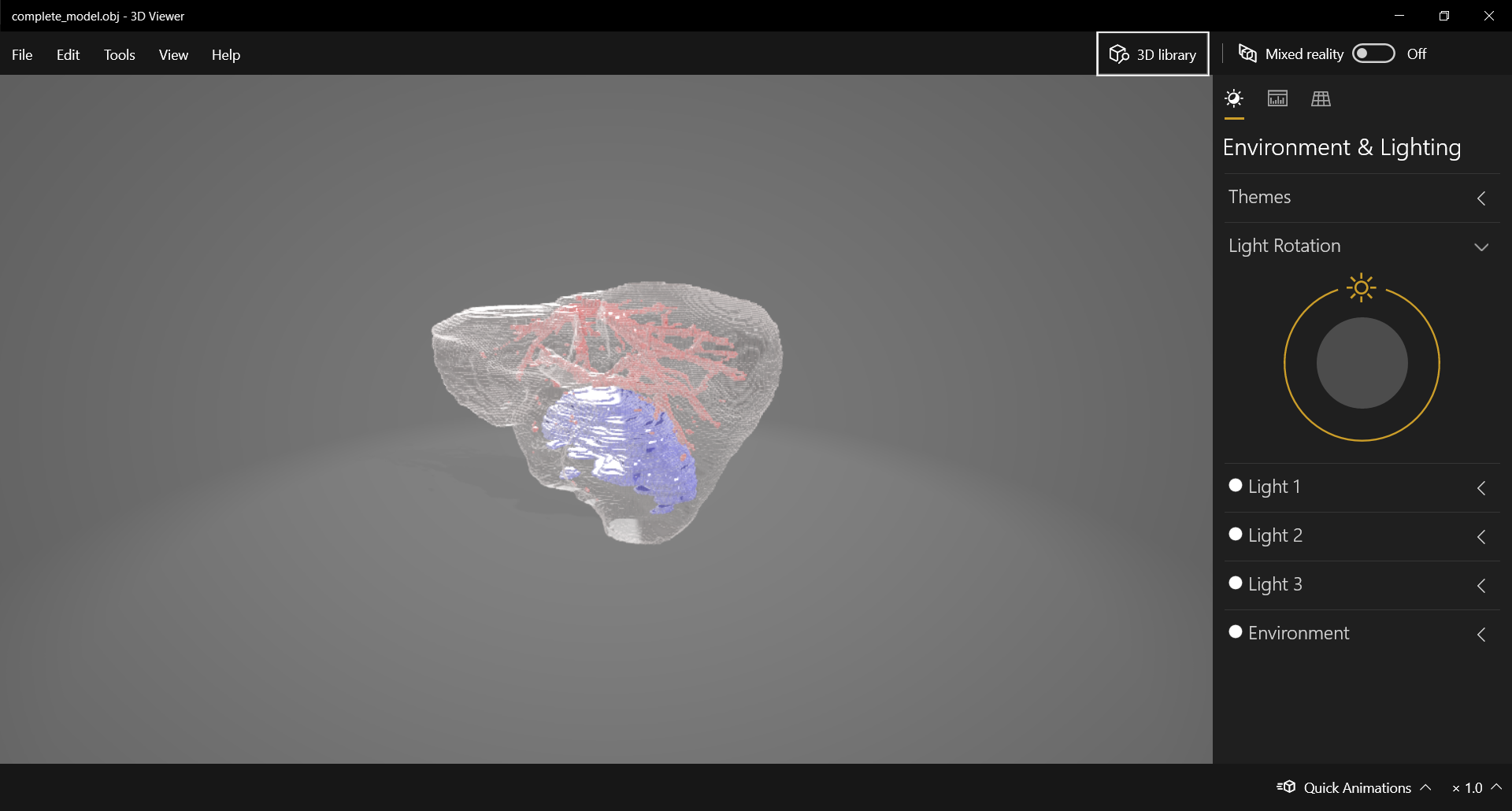}
    \caption{3D Paint software in Windows 10 viewing the constructed 3D object.}
    \label{fig:resulting_obj_mtl}
\end{figure}

The \enquote{complete\_model.obj} file comprises all vertices and faces needed to represent the volume in three dimensions, while the accompanying .mtl file assigns colors to distinguish among the three tissue types. Exporting these files in .obj and .mtl formats is particularly suitable for multi-colored 3D printing as suggested by Chakravorty~\cite{Chakravorty}. With access to a 3D printer, these liver volumes can be printed directly. Otherwise, they can be explored and analyzed through any software capable of rendering 3D models, as illustrated in Figure~\ref{fig:resulting_obj_mtl}.

\begin{figure}[!ht]
    \centering
    \includegraphics[width=1\textwidth]{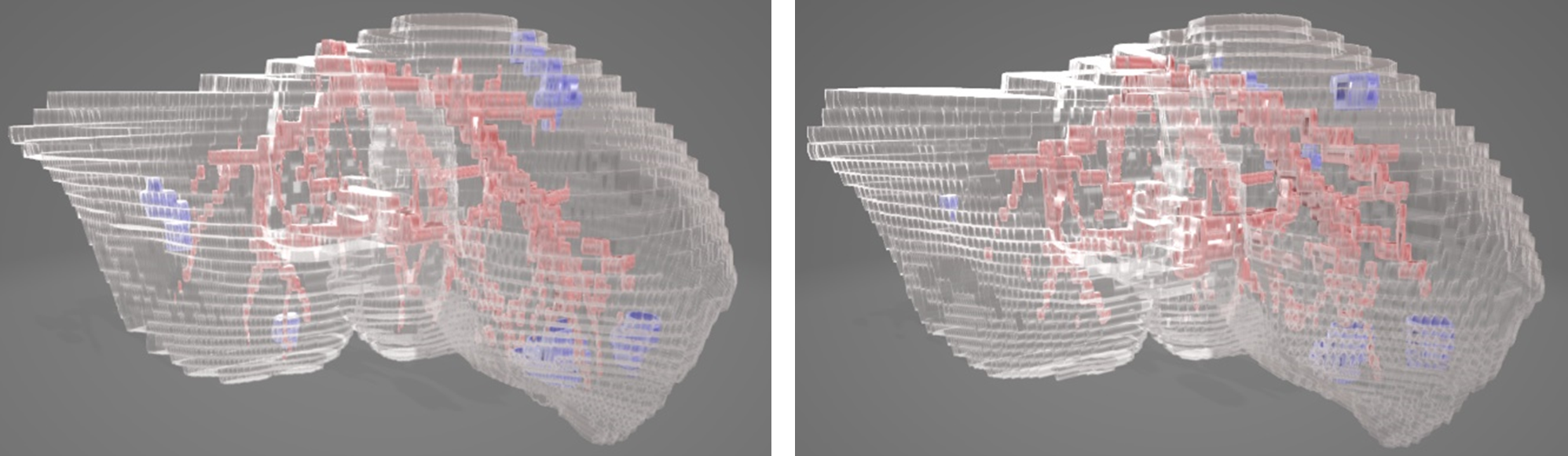}
    \begin{subfigure}[b]{0.49\textwidth}
    \caption{Ground-Truth}
    \label{fig:hepaticvessel-268-a}
    \end{subfigure}
    \hfill
    \begin{subfigure}[b]{0.49\textwidth}
    \caption{Our Segmentation}
    \label{fig:hepaticvessel-268-b}
    \end{subfigure}
    \caption{Volume hepaticvessel\_268 3D interpolation.}
    \label{fig:hepaticvessel-268}
\end{figure}

\begin{figure}[!ht]
    \centering
    \includegraphics[width=1\textwidth]{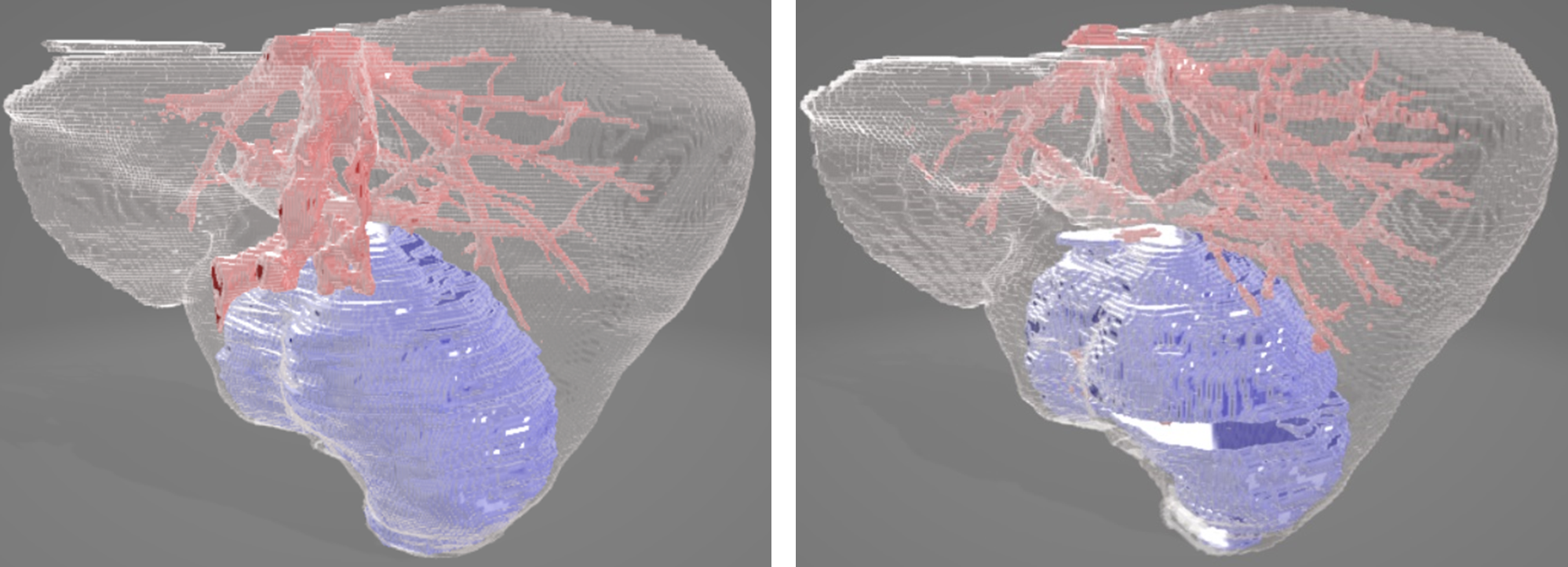}
    \begin{subfigure}[b]{0.49\textwidth}
    \caption{Ground-Truth}
    \label{fig:hepaticvessel-294-a}
    \end{subfigure}
    \hfill
    \begin{subfigure}[b]{0.49\textwidth}
    \caption{Our Segmentation}
    \label{fig:hepaticvessel-294-b}
    \end{subfigure}
    \caption{Volume hepaticvessel\_294 3D interpolation.}
    \label{fig:hepaticvessel-294}
\end{figure}

Coming back to the two test cases that have been shown in Figure~\ref{fig:2d-segmentation-showcase}, the following volumes in Figure~\ref{fig:hepaticvessel-268} and Figure~\ref{fig:hepaticvessel-294} show the constructed 3D objects through our AI Radiologist software, by utilizing the three developed ConvNets, and the 3D interpolation algorithm, all automatically completed in a matter of a minute! When we say the application is a real-time system, we mean by that that it would take significantly less time than manually annotating the volumes. It takes more than an hour to segment a liver, its tumors, and its vessels by an expert radiologist, while our system achieves similar precision within 2 minutes, depending on the volume's size (i.e., number of slices).

\section{Conclusions and Future Directions}\label{sec5}
To conclude, this paper targets developing a complete system from end-to-end named AI Radiologist, which takes care of developing deep ConvNets, employing them within an application, and prepareing the 3D printable objects in a matter of a minute. The developed methodology focuses primarily on applying preprocessing techniques over the CT scans and transforming them into 2.5D input to the liver segmentation ConvNet, which is based on the U-Net architecture. Initially, the output from the first model is employed to accurately locate and outline the liver, setting the stage for subsequent tumor and vessel ConvNets. The liver ConvNet demonstrates exceptional results, achieving an average Dice score of 98.12\% on a test set that includes 23 records from the MSDC-T8 dataset. Of these, 14 records provided ground-truth masks for tumors and vessels, with tumor segmentation achieving an average Dice score of 60\%, the top model reaching 65.95\%. Vessel segmentation, meanwhile, averaged around 50\% Dice score, with the highest-performing model scoring 51.94\%. Following segmentation, all masks are integrated and transformed into a 3D model using \enquote{The Marching Cubes} algorithm, which facilitates the generation of both .obj and .mtl files. To round off the process, the user-friendly AI Radiologist application, developed using the PyQt5 framework, deploys these advanced algorithms, assisting in the diagnosis of liver-related conditions. The application requires input in NIfTI format and outputs interpolated 3D models and segmented masks for each tissue type.

The inadvertent removal of vessels' voxels during the process of applying the liver mask to CT slices is a problem that requires attention. Additionally, our investigation into alternative ConvNet architectures was not exhaustive, which could have enriched our findings. Looking ahead, there are plans to incorporate DICOM files into our framework and to achieve beyond state-of-the-art results in both tumor and vessel segmentation. We aim to refine the localization strategy for training the vessel ConvNet using more effective methods. Furthermore, a specific case from the MSDC-T8 dataset (hepaticvessel\_072) consistently challenges the tumor ConvNets, leading to less than optimal performance. Addressing the complications posed by this particular scan could potentially enhance tumor ConvNet results by 4 – 5\%. Also on our agenda is improving the 3D visualization of the liver, particularly smoothing the surface of models derived from slices with low resolution, which currently exhibit almost staircase-like edges.

\section*{Acknowledgment}
This publication was made possible by an Award [GSRA6-2-0521-19034] from Qatar National Research Fund (a member of Qatar Foundation). The contents herein are solely the responsibility of the authors. Moreover, the HPC resources and services used in this work were provided by the Research Computing group in Texas A\&M University at Qatar. Research Computing is funded by the Qatar Foundation for Education, Science and Community Development (http://www.qf.org.qa).

\end{document}